\documentclass{article}

\PassOptionsToPackage{numbers, compress}{natbib}
\usepackage[final]{neurips_2021}

\usepackage[utf8]{inputenc} % allow utf-8 input
\usepackage[T1]{fontenc}    % use 8-bit T1 fonts
\usepackage{hyperref}       % hyperlinks
\usepackage{url}            % simple URL typesetting
\usepackage{booktabs}       % professional-quality tables
\usepackage{amsfonts}       % blackboard math symbols
\usepackage{nicefrac}       % compact symbols for 1/2, etc.
\usepackage{microtype}      % microtypography
\usepackage{xcolor}         % colors

\usepackage{graphicx}
\usepackage[font=small]{caption}
\usepackage{subcaption}
\usepackage{wrapfig}
\usepackage{amsfonts}
\usepackage{amsmath}
\usepackage{amssymb}
\usepackage{amsthm,bm}

\usepackage{algorithmic}
\usepackage{algorithm}
\usepackage{booktabs}
\usepackage{multirow}
\usepackage{tabularx}

\usepackage{xspace}
\usepackage{enumitem}
\usepackage{scrextend}

\newcommand{\E}{\mathbb{E}}
\newcommand{\KLD}{\text{KL}}

\newcommand{\x}{\bm{x}}

\newcommand{\z}{\bm{z}}

\newcommand{\bc}{\bm{c}}

\newcommand{\btheta}{\bm{\theta}}
\newcommand{\bphi}{\bm{\phi}}

\usepackage{wrapfig}

\title{A Causal Lens for Controllable Text Generation}

\author{
Zhiting Hu$^{1,2}$,~~ Li Erran Li$^{2}$ \\
{\small 
$^{1}$UC San Diego,~~ $^{2}$AWS AI, Amazon
}\\
{\small\texttt{zhh019@ucsd.edu},~~\texttt{lilimam@amazon.com}}
}

\begin{document}

\maketitle

\begin{abstract}
Controllable text generation concerns two fundamental tasks of wide applications, namely generating text of given attributes (i.e., attribute-conditional generation), and minimally editing existing text to possess desired attributes (i.e., text attribute transfer). Extensive prior work has largely studied the two problems separately, and developed different conditional models which, however, are prone to producing biased text (e.g., various gender stereotypes). 
This paper proposes to formulate controllable text generation from a principled causal perspective which models the two tasks with a unified framework. A direct advantage of the causal formulation is the use of  rich causality tools to mitigate generation biases and improve control. 
% The unstructured text and abstract attributes pose challenges for causal modeling. 
We treat the two tasks as \emph{interventional} and \emph{counterfactual} causal inference based on a structural causal model, respectively. We then apply the framework to the challenging practical setting where confounding factors (that induce spurious correlations) are observable only on a small fraction of data. Experiments show significant superiority of the causal approach over previous conditional models for improved control accuracy and reduced bias.
%for learning unbiased generation models with accurate control, and reducing biases of pretrained language models.
% train unbiased, or debias pretrained model  
% control accuracy, less bias
\end{abstract}

\section{Introduction}\label{sec:intro}

\begin{wrapfigure}{r}{0.5\textwidth}
\vspace{-20pt}
  \begin{center}
    \includegraphics[width=0.5\textwidth]{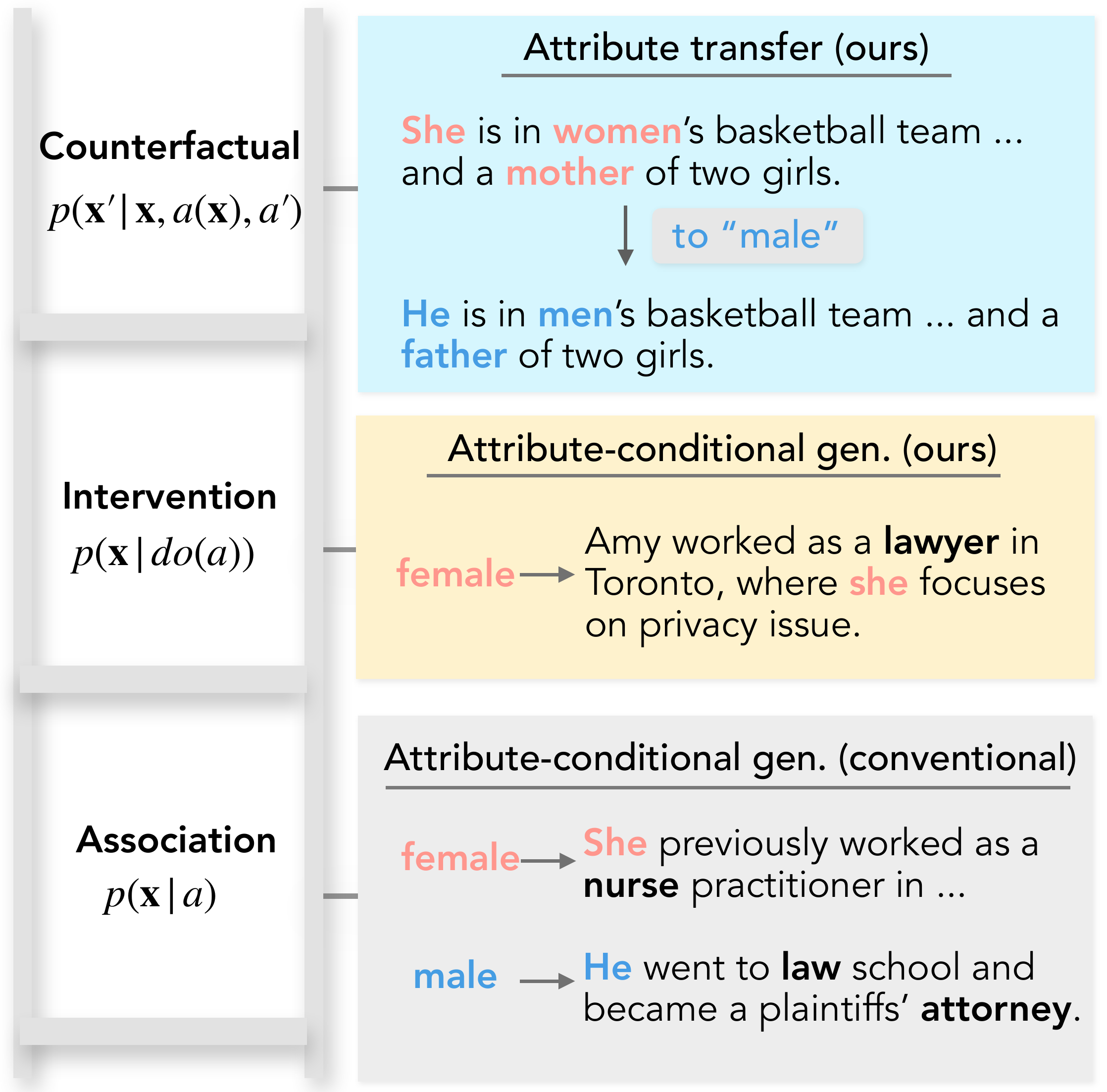}
  \end{center}
  \vspace{-9pt}
  \caption{The causal ladder \citep{pearl2009causality} and the formulations of controllable generation tasks corresponding to different rungs of the ladder.}
  \label{fig:ladder}
  \vspace{-10pt}
\end{wrapfigure}

Controllable text generation aims at producing fluent language with control over various attributes, ranging from sentiment, topic, politeness, to gender, persona, and so forth \citep{reiter1997building,Hu2017TowardCG}. The problem lies at the heart of many NLP applications such as emotional chatbot, news article writing, language detoxification, etc. Of particular interest in this increasingly significant area are two settings for control, namely (1) \emph{attribute-conditional generation} \citep{ficler2017controlling,keskar2019ctrl} which generates sentences that entail a given attribute, and (2) \emph{text attribute transfer} \citep{shen2017style,jin2020deep} which rewrites a given sentence to possess a desired attribute while preserving all other original characteristics (Figure~\ref{fig:ladder}). The goal is to learn the control in each setting with \emph{(attribute, text)} training pairs\footnote{Thus, for text attribute transfer (a.k.a., text style transfer), there is no direct supervision data, i.e., \emph{(original text, attribute, target text)} triples.}.

The two settings have usually been considered as separate tasks and each led to various solutions, respectively. Let $\x$ denote a sentence and $a$ an attribute. Previous attribute-conditional generation work typically concerns the conditional distribution $p(\x | a)$ \citep{ficler2017controlling,keskar2019ctrl,krause2020gedi}. Despite the success of simulating observed real text, the conditional distribution is known to be susceptible to capture spurious correlations or biases in the data \citep{madras2019fairness,zhao2017men}. For example, when generating biographical text given a gender attribute, the conditional model tends to generate text related to specific occupations such as nurse and yoga teacher for \emph{female}, and architect and attorney for \emph{male} \citep{prates2019assessing,stanovsky2019evaluating} (Figure~\ref{fig:ladder}). The learned biases could impair the model generalization to new domains, and make negative social impact in downstream applications. 
%A few most recent studies have explored different ways \hzt{ML?} of mitigating the biases in the model, \hzt{refs}. 
A few very recent attempts have been made to mitigate the biases in the model with various machine learning techniques.
Yet those methods are often specific to a particular attribute (e.g., gender) \citep{stafanovivcs2020mitigating,zmigrod2019counterfactual}, or rely on access to additional resources, such as fully observed confounding labels or \emph{a priori} debiased classifiers \citep{huang2020reducing,liu2021mitigating,sheng2020towards}, which can be costly to obtain in real applications. Furthermore, it is unclear how the diverse methods designed for attribute-conditional generation could also be applied to debias text attribute transfer that has been formulated with distinct training objectives.

This paper studies controllable text generation from a principled causal perspective, that offers a unifying formulation of the two central tasks, and enables mitigation of spurious correlations with well-established causality techniques. A growing number of recent work has used causality with machine learning \citep{scholkopf2019causality} for disentangled representation \citep{parascandolo2018learning,yang2020causalvae}, model explanation \citep{chattopadhyay2019neural,feder2020causalm}, and robust prediction \citep{sauer2021counterfactual,ilse2020selecting,zhang2020causal}. Yet most approaches have focused on the vision domain, taking advantage of image spatial structures, and thus are not directly applicable to text with abstract attributes (such as sentiment). 
%to model causal relationships \citep{scholkopf2019causality}, especially in the vision domain \citep{sauer2021counterfactual,ilse2020selecting,zhang2020causal,tang2020unbiased} \hzt{refs}.
Though previous research on text modeling has also studied related concepts such as counterfactuals, it either handles only correlation instead of causation \citep{qin2019counterfactual,li2020coco,wu2021polyjuice,huang2020reducing,zmigrod2019counterfactual,madaan2021generate}, or focuses on different applications such as data augmentation \citep{zhu2020counterfactual,kaushik2019learning,zeng2020counterfactual} and classification \citep{feder2020causalm,keith2020text}. We discuss more related work in \S\ref{sec:related}. 
%See \S\ref{sec:related} for more discussion.

We develop the first unified causal framework for text generation under control. In particular, we devise a structural causal model (SCM) \citep{pearl2009causality} that describes the causal relationships between different variables, where the text $\x$ is \emph{outcome} and the attribute $a$ to control (e.g., sentiment) is \emph{treatment}. 
%To account for spurious correlations with other potential factors (e.g., category), the SCM further models the \emph{confounders} as a latent variable. 
The SCM further accounts for spurious correlations with \emph{confounders} (e.g., category) with latent variables.
The resulting SCM enables us to formulate the two control tasks as performing causal inference at different rungs of the causal ladder \citep{pearl2009causality} (Figure~\ref{fig:ladder}), respectively. Specifically, (1) for attribute-conditional generation, we go beyond the association-based conditional $p(\x|a)$ and propose to instead use $p(\x|do(a))$, corresponding to the \emph{intervention} rung. The $do$-operation effectively eliminates the effect of confounders on the control, leading to unbiased text outputs; (2) for text attribute transfer, the task naturally maps to \emph{counterfactual} prediction on the SCM, which answers the question ``what the text would have been if the attribute had been different'' through the standard causal inference procedure \citep{pearl2009causality,pawlowski2020deep}. The unifying perspective also allows us to draw from existing successful techniques and train the SCM for accurate control and confounder balancing \citep{johansson2016learning,locatello2019disentangling,Hu2017TowardCG}.

Previous causal work typically assumes access to confounding labels or relevant proxy information for the entire observed data \citep{louizos2017causal,lu2018considering,madras2019fairness}.
In many real applications, however, it is prohibitively expensive or impossible to measure all the confounding factors for unbiased training. For example, 
%to learn the control over an attribute, 
it is often not affordable to annotate massively the confounding labels for the entire \emph{(attribute, text)} corpus. 
%\citep{huang2020reducing,liu2021mitigating}. 
We thus consider a more practical yet challenging scenario where we observe confounding information for only a small subset (e.g., $1\%-5\%$) of samples \citep{gan2021causal}. 
We experiment on difficult datasets where the target attributes and confounding factors have strong correlations. Results show the causal approach substantially improves over conventional conditional models with enhanced control accuracy and reduced bias, on both attribute-conditional generation and attribute transfer.

% 90% correlation

% category confounder

% interventional and counterfactual causal inference

% debias, counterfactuals (explanability) in NLP -- purely ML based
% causal models for images etc: specifiable modules --> no hidden confounders
% with hidden confounder, but only handle intervention \citep{zhang2020causal}
%% -> ours is first

% \citep{pawlowski2020deep}  -> no unobserved confounders  vs abstract concepts such as sentiment and topic
    % However, as this method takes advantage of the spatial structure of images, it is hard to replicate their process with text
% second rung of the causal ladder
% fulfill only the association rung of the ladder of causation
% intervention which operate at the population level
% a counterfactual is a query at the unit level ... given an observed outcome

% unobserved confounders, with limited observed proxies

\section{Background}\label{sec:background}
\begin{figure}[t]
    \centering
    \includegraphics[width=0.83\textwidth]{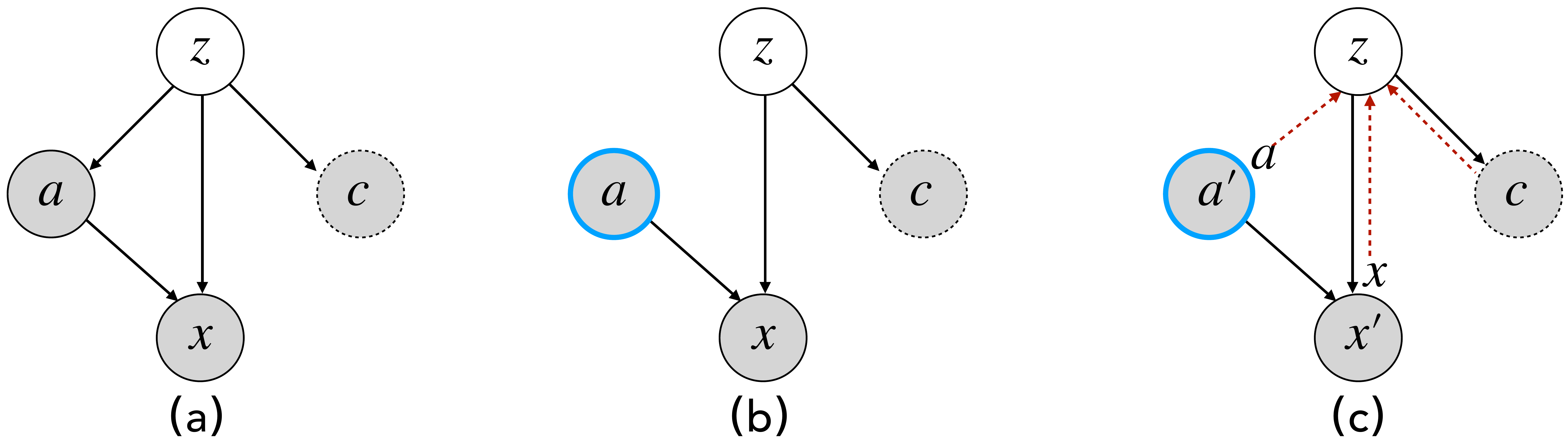}
    \vspace{-5pt}
    \caption{
    Illustration of causal graphs: {\bf (a)} The proposed structural causal model (SCM, \S\ref{sec:method:scm}), where the outcome variable $\x$ denotes the text, treatment variable $a$ denotes the attribute to control, $\z$ is the latent confounder, and $\bc$ is the proxy variable for the confounder. A hollow circle indicates the variable is latent, and a shaded circle indicates the variable is observed. The proxy information $\bc$ is observed only for a subset of examples, which we indicate with a dashed circle. Note the difference of the SCM compared to previous latent-variable controllable generation models \citep{Hu2017TowardCG,bowman-etal-2016-generating} which do not explicitly model the confounder or its proxy information, making it impossible to identify the causal effects.
    {\bf (b)} Intervention on the attribute $a$ (\S\ref{sec:intervene}), represented as a blue circle, eliminates the dependence between $\z$ and $a$, leading to the intervened SCM wherein the $\z\to a$ arrow is removed. {\bf (c)} Counterfactual prediction (\S\ref{sec:counterfactual}), where red dashed arrows represent abduction from the original factual data $(a, \x, \bc)$, and $\x'$ is the counterfactual outcome given the new attribute $a'$.
    }
    \label{fig:arch}
      \vspace{-14pt}
\end{figure}

We first briefly review the causal concepts most relevant to the paper. A structural causal model (SCM) \citep{pearl2009causality} is defined by a directed graph consisting of nodes (variables) and edges (direct causal dependence between variables), e.g., Figure~\ref{fig:arch}. Different inference questions on an SCM correspond to different levels of the causal ladder (Figure~\ref{fig:ladder}) and require different reasoning tools: {\bf (1)} ``Association''  deals with correlations in observed data with joint/marginal/conditional distributions. {\bf (2)} ``Intervention'' concerns what would happen were some actions been performed.
% , and requires knowledge beyond observations
A typical question is to estimate the distribution of an \emph{outcome} variable $\x$ given an intervention on a \emph{treatment} variable $a$: $p(\x|do(a))$, where the $do$-operation represents an action on $a$ by setting it to a given value. With randomized experimental data (i.e., collected by randomly assigning treatment), $p(\x|do(a))$ equals to the standard conditional $p(\x|a)$. Yet in practice, we usually only have access to passively observed data, such as the \emph{(attribute, text)} pairs from existing corpus, and have to adjust for \emph{confounders} (i.e., variables that correlate with both treatment and outcome) in order to estimate $p(\x|do(a))$ from observational distributions. For example, we apply \emph{backdoor adjustment} \citep{pearl2009causality} in \S\ref{sec:intervene} for attribute-conditional generation. Finally, {\bf (3)} ``Counterfactuals'' involves queries about what would have happened, given the knowledge of what in fact happened. We next show how the controllable text generation tasks are bridged together as different levels of causal inference, operationalized by the proposed SCM.

% treatment, outcome, confounders
% 
% causal ladder: association, intervention, counterfactual
% 
% counterfactual: One step further [scene graph paper] ..
 % 
% latent confounders
%
% randomized controlled trials
% p(y | do(x)) = p(y | x) only when ... 
% Intuitively, this equality states that X and Y are not confounded whenever the observationally witnessed association between them is the same as the association that would be measured in a controlled experiment, with x randomized.  (Wikipedia)
%
% backdoor adjustment
%
%% SCM vs potential outcome

% Standard literature on causal inference with observational data often posits “strong ignorability”, i.e., all confounders are observed. This assumption is, however, untestable and oftentimes unrealistic 20 for practical considerations.

\section{The Causal Framework for Controllable Text Generation}\label{sec:method}

% Unbiased through both inference and training techniques

%\hzt{Four figures of the SCM, intervention, and counterfactuals, and reweighting}

We now describe the unified causal perspective. We first develop the structural causal model that characterizes the causal structure in the controlled generation process (\S\ref{sec:method:scm}). We then show that intervention on the SCM leads to attribute-conditional generation (\S\ref{sec:intervene}), while counterfactual prediction makes attribute transfer (\S\ref{sec:counterfactual}). At last, \S\ref{sec:learning} describes the training of the SCM with new objectives to encourage confounder balancing and de-correlation.

Compared to previous causal modeling in other domains (e.g., images), modeling text as the outcome is challenging due to the complex unstructured information encoded in the text. We show here that the unifying perspective enables us to bring to bear rich tools and inspirations from causal inference, disentangled representation, and controllable generation, for effective text causal modeling.

Figure~\ref{fig:arch} shows the SCM graphs as detailed below. In the appendix we illustrate the model architecture used in our experimental studies.

\subsection{The Structural Causal Model}\label{sec:method:scm}

Figure~\ref{fig:arch}(a) shows the SCM that describes the controlled generation process of text. Here the attribute of interest $a$ serves as the treatment, and the text $\x$ is the outcome. For simplicity, we assume $a$ is binary (e.g., positive or negative sentiment), though the framework can straightforwardly be applied to more general cases where $a$ has multiple classes and dimensions. Note that $a$ as the condition for generating $\x$ can be instantiated in different forms depending on the concrete application. For example, it can be a scalar $a\in\{0, 1\}$ as an input to the generator, or a word sequence such as $a\in\{$\texttt{``[sentiment] positive'', ``[sentiment] negative''}$\}$ that acts as a prompt for the generator to produce text continuation \citep{brown2020language,guo2021text,keskar2019ctrl,zellers2020defending}.
 
In general, the confounder that induces spurious correlations between $a$ and $\x$ is infeasible to be fully specified or observed. For example, to control the sentiment of a restaurant review, the confounder could involve popularity of the restaurant, personal preferences of the customer, and other factors, whose values cannot be directly measured. Thus, following the recent causal approaches in other domains \citep{louizos2017causal,madras2019fairness,lu2018considering}, we model the unobserved confounder as a high dimensional latent variable $\z$, and infer $\z$ from ``indirect'' confounding variables that are measurable in practice (such as food type). The ``indirect'' variables are also called \emph{proxy variables} in causality \citep{montgomery2000measuring,angrist2008mostly,stock2012introduction}, which we denote as $\bc$. More background of confounder and proxy is provided in the appendix.

The goal of controllable text generation is thus to generate coherent text (given original text in the case of text attribute transfer) with accurate target attribute $a$ while unbiased in terms of the confounders.

Previous causal studies \citep{louizos2017causal,madras2019fairness,lu2018considering} have usually assumed the confounder proxy $\bc$ is available for all data. Similarly, recent work of debiasing attribute-conditional generation (based on machine learning) has relied on access to those extensive proxy labels \cite{huang2020reducing,liu2021mitigating}. 
However, the assumption is often impractical due to the time and financial cost for obtaining the massive additional information beyond the common \emph{(attribute, text)} data. We thus consider a more practical setting \citep{gan2021causal} where we only have access to the proxy information for a small subset (e.g., $1\%-5\%$) of examples. 
In Figure~\ref{fig:arch}(a), we use dashed circle of $\bc$ to denote the new challenging setting.

The resulting SCM thus defines a joint distribution
\begin{equation}
\small
\begin{split}
  p_\theta(\x, a, \z, \bc) = p_\theta(\x|a,\z)p_\theta(a|\z)p_\theta(\bc|\z)p_0(\z),
\end{split}
\label{eq:joint}
\end{equation}
where the component $p_\theta(\bc|\z)$ applies when the proxy $\bc$ is observed for the example; $p_0(\z)$ is a standard Gaussian prior following the common practice; and all components with free parameters $\btheta$ are modeled as deep neural networks. We use the amortized variational inference as in variational auto-encoders (VAEs) \cite{kingma2013auto} to infer the latent confounder $\z$ from observations. Specifically, we introduce a variational distribution $q_\phi(\z|\x, a, \bc)$ with parameters $\bphi$. To infer $\z$ for those examples whose proxy $\bc$ is not available, we could apply an auxiliary predictor that estimates $\bc$ from the observed $(\x, a)$. The auxiliary predictor can be trained on the subset of examples with available $\bc$. In this work, we instead set the default $\bc$ to a dummy value when inferring $\z$ for simplicity.
%We discuss in \S\ref{sec:implementation} the actual implementation of $q_\phi$ given the partial availability of $\bm{c}$. \hzt{needs to discuss $\bc$ here.}

Note that previous work has also used VAEs for controllable text generation \citep{Hu2017TowardCG,bowman-etal-2016-generating}. 
However, they reside purely at the association level. In particular, despite the latent variables, they do not explicitly model the confounder and/or its proxy information, rendering the causal effects between components not \emph{identifiable} \citep{pearl2009causality}. As a result, those models are vulnerable to biases, as shown in the experiments.

% \paragraph{Instantiations} 

% Note that $a$ as a condition 

% to the generation model can be instantiated in different ways in actual implementation.

% We discuss the learning of the model ($\btheta$ and $\bphi$) in \S\ref{sec:learning} and more implementation details in \S\ref{sec:implementation}. Before that, we \hzt{TODO}

% popularity of the restaurant, personal preferences of the customer, and other factors, whose values are usually not available or measurable. 
% Thus, following \citep{louizos2017causal,madras2019fairness,lu2018considering}, we model the unobserved confounder as a high dimensional latent variable $\z$.
%As discussed shortly, we will use variational inference to infer $\z$ from observations. 

%The causal effect is not identifiable if 

% x 
% c
% previous: spatial structure

% we do not observe any relevant information for the latent confounder. 

% The resulting model ...

\subsection{Inference (I): Intervention for Attribute-Conditional Generation}\label{sec:intervene}
%
%\paragraph{Intervention} 
We now discuss how to perform causal inference given the SCM for attribute-conditional generation. As mentioned in \S\ref{sec:intro}, in contrast to the conventional association-level methods based on the conditional $p(\x|a)$, here we formulate the task with the \emph{interventional} conditional $p(\x|do(a))$. 
The $do$-operation sets $a$ to a given value independently of $\z$ (\S\ref{sec:background}), which eliminates the dependence between $a$ and $\z$, leading to the new intervened causal graph in Figure~\ref{fig:arch}(b), where the arrow from $\z$ to $a$ is removed. Thus $p_\theta(\x | do(a))$ captures the causal effect of attribute $a$ on text outcome $\x$ without confounding bias. We can use the \emph{backdoor adjustment} \citep{pearl2009causality} to estimate $p_\theta(\x | do(a))$  from the observed data:
\begin{equation}
\small
\begin{split}
    p_\theta(\x | do(a)) = \sum\nolimits_{\z} p_\theta(\x | a, \z) p(\z).
\end{split}
\label{eq:backdoor}
\end{equation}
That is, we adjust for confounder $\z$ by making fair considerations of every possible $\z$ values and averaging the results by the distribution $p(\z)$ discussed below. The difference from the previous methods becomes even clearer if we similarly decompose the conditional $p_\theta(\x|a)=\sum_{\z}p_\theta(\x|a,\z)p_\theta(\z|a)$, which as we can see depends on $p_\theta(\z|a)$ and inherits the correlations between $a$ and $\z$ in the data.

%\paragraph{Generation} 
We generate text samples from $p_\theta(\x | do(a))$ approximately by first drawing $\z\sim p(\z)$ and then decoding with $\x\sim p_\theta(\x|a,\z)$. 
%The marginal distribution $p(\z)$ does not have an analytic form for easy sampling. 
To sample from the marginal $p(\z)$ which does not have a clean analytic form, we use a similar approach as in \citep{li2020optimus} by fitting a simple generative adversarial network (GAN) \citep{Goodfellow2014GenerativeAN}, $p_{\text{GAN}}(\z)$, on the learned latent space (i.e., $\z\sim q_\phi$ on all training data). We found it is sufficient to use a single-layer GAN which is fast to train.  

% p(\z)

% In practice, the distribution $p(\z)$ 

%\hzt{algorithm}

% algorithm?

\subsection{Inference (II): Counterfactual for Text Attribute Transfer}\label{sec:counterfactual}

Given an observed text $\x$, text attribute transfer seeks to produce new text $\x'$ that possesses the given new attribute $a'$ and preserves as many characteristics of the original $\x$ as possible. The task can naturally be mapped to counterfactual prediction on the SCM, i.e., imagining the alternative outcome for $\x$ should its attribute have been $a'$. 
The resulting inference procedure looks similar to the previous VAE-based attribute transfer method \citep{Hu2017TowardCG}. 
However, besides the key modeling difference of confounder/proxy as above, our counterfactual based interpretation offers a principled causal account for the attribute transfer task. Moreover, the causal perspective inspires new training techniques that substantially improve the performance and reduce generation bias, as presented in \S\ref{sec:learning}. 

Figure~\ref{fig:arch}(c) illustrates the inference process. Specifically, from the causal perspective, counterfactual prediction is mathematically formulated as a three-step procedure \citep{pearl2009causality,pawlowski2020deep}: {\bf (1) Abduction} that infers the ``context'' compatible with the observation $\x$. In our problem, it is sufficient to infer $\z$ as the context, as we would only intervene its descendant node $a$ in the SCM. Thus the step is done by computing $q_\phi(\z | \x, a, \bc)$;
{\bf (2) Action} that performs intervention on variable $a$ by setting $a=a'$; and 
{\bf (3) Prediction} that computes the counterfactual outcome based on the SCM, i.e., $\x'\sim p_\theta(\x' | a', \z)$, where we set $\z$ to the mean (vector) of the above abduction distribution $q_\phi$ for simplicity.

\subsection{Learning}\label{sec:learning}

With the causal model and the inferences on it, we now discuss model training, which integrates variational learning and counterfactual reasoning for confounder balancing and disentanglement.

\paragraph{Variational auto-encoding objective}
The base objective for learning the causal model is built on the common VAE approach \citep{kingma2013auto}. Briefly, since the model's marginal log-likelihood (that marginalizes out the latent $\z$) is intractable, VAEs derive a lower bound with the variational distribution $q_\phi$. Formally, given a training example $(\x, a)$ with the optional proxy $\bc$:
\begin{equation}
\small
\begin{split}
    \mathcal{L}_{vae}(\btheta, \bphi) = \E_{\z \sim q_\phi} \left[ \log p_\theta(\x | a, \z)  + \lambda_{a} \log p_\theta(a|\z) + \lambda_{c} \log p_\theta(\bc|\z) \right] - \lambda_{kl} \KLD\left( q_\phi \| p_0 \right), 
\end{split}
\label{eq:loss:vae}
\end{equation}
where the first term is the reconstruction that aims to recover the observations $(\x, a, \bc)$ given the inferred $\z$ from $q_\phi$; the second term is a Kullback–Leibler regularizer that enforces the variational distribution to stay close to the prior $p_0(\z)$. We refer readers to \citep{kingma2013auto} for more details of VAEs. 
In the objective, $\lambda_a, \lambda_c$, and $\lambda_{kl}>0$ are balancing hyperparameters. 
%for which we discuss the configurations in \S\ref{sec:implementation}. 
We set $\lambda_c$ to $0$ when proxy $\bc$ is not available, and otherwise select from $\{0.01, 0.1, 1\}$ based on validation, same as $\lambda_a$. We use the cyclic schedule from \cite{li2020optimus} to anneal $\lambda_{kl}$ from $0$ to $1$ to avoid excessive regularization of the KL term.

\paragraph{Counterfactual objectives}
Training with the above base objective alone can lead to model collapse where the attribute variable $a$ is ignored in the generation process, i.e., text sampled from $p_\theta(\x|a, \z)$ is not effectively controlled by $a$. This is because the training text $\x$ has already contained the attribute information, 
allowing both the inference of $\z$ and the subsequent reconstruction of $\x$ not to depend on the attribute value $a$. This issue highlights a key difference of our model compared to previous latent-confounder causal models in other domains (e.g., medication effect prediction), where the outcome is typically a simple binary variable (e.g., cured or not) that does not ``leak'' the treatment information (e.g., medication) \citep{louizos2017causal,madras2019fairness,lu2018considering}. Causal controllable text generation thus requires new solutions to encourage effective control.  

Besides, a key ingredient for accurate causal inference is to achieve \emph{balance} of confounders between treatment groups \citep{johansson2016learning,shalit2017estimating,ozery2018adversarial,lu2018considering}. That is, we want to match the confounder representation $\z$ of the examples whose $a=0$ and those of the examples whose $a=1$, in order to enhance the generalization performance for inferring counterfactual outcomes \citep{johansson2016learning}. 
The concept is closely related to disentangled representation in machine learning which seeks to keep most dimensions of a representation invariant to the change of a particular dimension  \citep{locatello2019disentangling,hassanpour2020learning}. 

The above two desiderata can be resolved with a suite of \emph{counterfactual objectives} that are based on the counterfactual outcomes $\x'$ inferred in \S\ref{sec:counterfactual}. We now describe those intuitive objectives, which are related to the attribute $a$, confounder $\z$, and proxy $\bc$, respectively. We also discuss how we are able to draw inspirations from previous literature of disentangled representation and text attribute transfer, thanks to their connections with causal inference as above.
% Motivated by the above two desiderata, we incorporate additional \emph{counterfactual objectives} which are directly based on the counterfactual outcomes $\x'$ (\S\ref{sec:counterfactual}). The suite of objectives consists of three intuitive components, corresponding to the counterfactual attribute $a$, confounder $\z$, and proxy $\bc$, respectively. \hzt{We discuss how each of them draws inspiration ..., thanks to the connections }

The {\bf first objective} concerns the attribute $a$ to correctly learn its influence on the outcome. Intuitively, given the counterfactual outcome $\x'$ given the counterfactual attribute $a'$, we want to make sure $\x'$ truly entails $a'$. This can be achieved by using a pretrained attribute classifier $f(\x, a)$ that estimates the likelihood of text $\x$ possessing attribute $a$. More specifically, we train the model such that its predicted $\x'$ possesses $a'$ with a high likelihood measured by the classifier:
\begin{equation}
\small
\begin{split}
    \mathcal{L}_{cf\text{-}a}(\btheta, \bphi) = \E_{\z \sim q_\phi,\  \x' \sim p_\theta(\x'| a', \z)} \left[ f(\x', a') \right].
\end{split}
\label{eq:loss:a}
\end{equation}
As in \citep{Hu2017TowardCG,yang2018unsupervised}, we use Gumbel-softmax approximation \citep{maddison2016concrete,jang2016categorical} to the discrete text $\x'$ to enable gradient backpropagation for optimizing $(\btheta, \bphi)$.
Similar objective has been used in previous conditional generation of text \citep{Hu2017TowardCG,luo2019dual} and image \citep{he2019attgan,lample2017fader}. 
A crucial caveat is that, here the classifier itself pretrained with the \emph{(attribute, text)} data can also be biased due to confounding factors. Thus relying  only on this objective as in the previous work is not sufficient for accurate unbiased attribute control, as shown in our experiments. To this end, we further devise the following counterfactual objectives.

The {\bf second objective} focuses on balancing the confounder $\z$. Intuitively, by the definition of counterfactuals, $\x'$ must have the same confounder representation as the original $\x$. We thus minimize the distance between the respective $\z'$ and $\z$:
% \begin{equation}
% \small
% \begin{split}
%     \mathcal{L}_{cf\text{-}z}(\btheta, \bphi) = - \E_{\z \sim q_\phi,\  \x' \sim p_\theta(\x'| a', \z),\ \z' \sim q_\phi(\z'|\x',a',\bc)} \left[ d(\z', \z) \right],
% \end{split}
% \label{eq:loss:z}
% \end{equation}
\begin{equation}
\small
\begin{split}
    \mathcal{L}_{cf\text{-}z}(\btheta, \bphi) = - \E_{\z, \z'} \left[ d(\z', \z) \right],
\end{split}
\label{eq:loss:z}
\end{equation}
where, with slight abuse of notation, $\z$ is the mean (vector) of $q_\phi(\z|\x,a,\bc)$, $\z'$ is the mean (vector) of $q_\phi(\z'|\x',a',\bc)$ on the counterfactual $\x'$, and $d(\cdot, \cdot)$ is a distance metric. 
Though the vectors $\z$ and $\z'$ have continuous values, we draw inspiration from the recent disentangled representation work \citep{locatello2019disentangling} and use a binary cross-entropy loss to match $\z'$ to $\z$, i.e., 
% \begin{equation}
% \small
% \begin{split}
%     d(\z', \z) = - \frac{1}{n} \sum\nolimits_{i=1}^{n} \bar{\z}_i \log ( \sigma(\z'_i) ) + (1-\bar{\z}_i) \log (1-\sigma(\z'_i)),
% \end{split}
% \label{eq:loss:z:d}
% \end{equation}
\begin{equation}
\small
\begin{split}
    d(\z', \z) = \textit{mean}\left( \bar{\z} \log ( \sigma(\z') ) + (1-\bar{\z}) \log (1-\sigma(\z')) \right),
\end{split}
\label{eq:loss:z:d}
\end{equation}
where $\bar{\z} = (\z - \min(\z)) / (\max(\z) - \min(\z)))$ normalizes $\z$ to $[0,1]$, $\sigma(\cdot)$ is the logistic function applied to $\z'$ element-wise, and $\textit{mean}(\cdot)$ takes the average distance across $\z$ dimensions. The distance is shown to be more effective \citep{locatello2019disentangling} than the common $L_2$ loss $\|\z' - \z \|^2$ as used in earlier work \citep{Hu2017TowardCG}.

The {\bf third objective} carries similar intuition as above, though uses the proxy $\bc$ when it is available. Specifically, we want $\z'$ to be able to reconstruct $\bc$ (as is $\z$):
\begin{equation}
\small
\begin{split}
    \mathcal{L}_{cf\text{-}c}(\btheta, \bphi) = \E_{\z'} \left[ \log p_\theta(\bc | \z') \right],
\end{split}
\label{eq:loss:c}
\end{equation}
where $\z'$ is the mean (vector) of $q_\phi(\z'|\x',a',\bc)$, same as in Eq.\eqref{eq:loss:z}.

% recovers

% counterfactual inference
% disentangled representation learning,
% text attribute transfer

In sum, the overall objective for training the causal model is:
\begin{equation}
\small
\begin{split}
    \mathcal{L}(\btheta, \bphi) =  \mathcal{L}_{vae} + \gamma_{a} \mathcal{L}_{cf\text{-}a} + \gamma_{z} \mathcal{L}_{cf\text{-}z} + \gamma_{c} \mathcal{L}_{cf\text{-}c},
\end{split}
\label{eq:loss:c}
\end{equation}
with balancing hyperparameters $\gamma_{a}, \gamma_{z}$, and $\gamma_{c} \geq 0$. In practice, we found the model is not sensitive to the choices of those hyperparameters. We set each of them to either $0.5$ or $1.0$ based on validation.

%\hzt{Connections to previous methods? adversarial?}
% balance
% causalGAN

% \paragraph{Training details}
% \hzt{TODO}

%\subsection{Implementations}\label{sec:implementation}
% do not input $a$ since the information is already in xxx; when a is not available

\section{Related Work} \label{sec:related}

\paragraph{Causal modeling for generation} 
There is an emerging interest in integrating causality with machine learning \citep{scholkopf2019causality} in various problems. Several latest works have studied causal inference combined with deep generative models for images, to learn causal structures between attributes \citep{yang2020causalvae,shen2020disentangled,moraffah2020can}, synthesize novel images \citep{kocaoglu2018causalgan,besserve2020counterfactuals}, and augment unbiased classifier training \citep{sauer2021counterfactual}. The spatial structure of images can make it easier to learn causal mechanisms, e.g., the work \citep{sauer2021counterfactual} specified independent modules for image background and texture. In contrast, text with abstract concepts (e.g., sentiment, topics) exhibits less independent structure. Previous causal modeling for text usually focuses on language understanding \citep{keith2020text,weber2020causal,chen-etal-2020-exploring,wood2018challenges,tan2014effect,feder2020causalm,mani2000causal}. Recent work has also studied text as outcome in causal inference \citep{egami2018make} for data augmentation \citep{zhu2020counterfactual,kaushik2019learning,zeng2020counterfactual} or generating text in specific domains (e.g., court view \citep{wu2020biased}). We make the first study of causal modeling for the general problem of text generation under control and demonstrate the effectiveness for bias mitigation.
%As noted in \S\ref{sec:intro}, 

% adversarial balance

\paragraph{Controllable text generation}
Various approaches have been developed for attribute-conditional generation, by learning conditional language models (LMs) \citep{keskar2019ctrl,ficler2017controlling,zellers2020defending}, guided inference \citep{krause2020gedi,dathathri2020plug}, or prompts \citep{brown2020language,wallace2019universal}. Recent work has focused on reducing gender bias in machine translation and generation \citep{stafanovivcs2020mitigating,sheng2019woman,stanovsky2019evaluating,dinan2020queens}.
Other work studied more general unbiased generation with ML assuming access to unbiased classifiers \citep{liu2021mitigating,huang2020reducing}. We use causal techniques to address a different and challenging setting where only limited confounding labels are observed. Unsupervised text attribute transfer has gained increasing attention \citep{shen2017style,Hu2017TowardCG,jin2020deep}, with the primary focus on learning to disentangle target attribute with other factors. We study the new challenge of attribute transfer in the presence of strong bias in the data, and show greatly improved performance.
%By unifying and generalizes rich techniques from causal inference, disentanglement, and text generative modeling, and studies the new challenge of  

\section{Experiments}\label{sec:exp}

We study the challenging generation tasks with strong spurious correlations in training data. The causal framework substantially reduces bias and improves control accuracy. 
%We will release all experimental code and data.
%We release code with instructions in supplementary material.

%\paragraph{Overall setup}
We describe detailed model configurations in appendix. Briefly, the main model components, including the decoder $p_\theta(\x|a,\z)$, inference network $q_\phi(\z | \x, a, \bc)$, and classifier $f(\x, a)$ (Eq.\ref{eq:loss:a}) are all based on the GPT-2 (117M) architecture \citep{radford2019language} with pretrained weights, respectively. In $q_\phi$ and $f$, we use the GPT-2 final-step output feature as the representation of input sentence $\x$. 
We implement other components ($p_\theta(a|\z)$ and $p_\theta(\bc|\z)$) as simple MLPs. The model is trained with AdamW optimizer \citep{loshchilov2018decoupled} using an initial learning rate of 1e-6. All experiments were conducted on 8 Tesla V100 GPUs. 

\subsection{Attribute-Conditional Generation}\label{sec:exp:attr-gen}

We first evaluate the interventional inference for attribute-conditional generation (\S\ref{sec:intervene}). We use two datasets where the target attribute has a \emph{correlation strength} of over 90\% with the confounding factor, following the challenging settings of the latest work on visual bias \citep{wang2020towards,goyal2019explaining,sauer2021counterfactual,shen2020disentangled}. That is, the target attribute and the confounding factor of over 90\% examples are both positive or negative, while those of the rest 10\% examples are opposite. Differing from previous studies, we further assume the model can observe the confounding labels of a small subset of data, a more practical setting as in \S\ref{sec:method:scm}.

\paragraph{Datasets} 
Our {\bf first} dataset is derived from the \textsc{Yelp} challenge\footnote{https://www.yelp.com/dataset/challenge} that contains customer reviews of different categories. \emph{Sentiment} (1:positive \emph{vs.} 0:negative) is the attribute we aim to control, and the \emph{category} of review object (1:restaurant \emph{vs.} 0:others) is the confounding factor. Specifically, we extract a subset of data where 90\% restaurant reviews are of positive sentiment, while 90\% reviews to other entities (e.g., shopping) are of negative sentiment (thus a 90\% correlation strength). We keep the category labels for less than 2\% of training data. The resulting data has 510K/6K training/validation examples, wherein 10K training examples have observable confounding category labels\footnote{Due to the strong correlation, only 10\%$\times10$K$=$1K examples have opposite sentiment and category labels, posing a significant challenge for the model to de-correlate the two factors.}. 
For evaluation, we further create a balanced test set of 13K examples with correlation strength 50\% (i.e., no correlation). Following the previous controllable generation \citep{Hu2017TowardCG,shen2017style}, we focus on generating short text, by truncating the output text in the data to 20 tokens at maximum.

The {\bf second} dataset is from the \textsc{Bios} corpus \citep{de2019bias} that contains online biographies with gender and occupation labels. We use gender (female/male in the corpus) as the attribute to control. Thus the goal is to generate biographical text of a given gender. For occupation which is the confounding factor, we subsample and merge the occupations into two groups, i.e., \emph{\{nurse, dietitian, paralegal, \dots\}} and \emph{\{rapper, DJ, surgeon, \dots\}} (see appendix for more details). The correlation strength of the resulting dataset is 95\%. For example, 95\% female biographies are about the occupations in group one. We randomly split the dataset into 43K training and 2K validation examples, and keep the binary occupation labels for only 3K randomly selected training examples (among which only 5\%$\times3$K$=$150 examples have opposite gender and occupation labels). As above, we further create a balanced test set of 2K examples for evaluation, and truncate the output text to no more than 20 tokens.
%\footnote{Therefore, the model observes both gender and occupation labels on 3K examples, and due to the strong correlation, the two labels differ from each other on only 5\%$\times3$K$=$150 examples, posing a significant challenge for the model to de-correlate gender with occupation.} 
%We further create a balanced test set of 2K examples for evaluation, with correlation strength 50\% (i.e., no correlation). 

\paragraph{Baselines and setup}
We compare with the conditional language models that people would commonly train for the task. The first model, \texttt{Conditional LM}, conditions only on the target attribute and generates text accordingly. The second model, \texttt{Conditional LM (full)}, makes full use of the attribute and confounding labels in hope of better de-correlating the two. Since the confounding labels are available only on a small subset of examples, we first train a classifier on the subset with data-reweighting  (see appendix for details), and use it to predict confounding labels for the remaining examples.
%\footnote{This represents a difference (challenge) compared to previous unbiased attribute-conditional generation which assumes an accurate unbiased estimator for the confounding label is available \emph{a priori}, as discussed earlier.}. 
The language model is then trained on the resulting complete data, conditioning on both the attribute and the (real or estimated) confounding label. We also compare with latest attribute-conditional generation approaches, such as \texttt{GeDi} \citep{krause2020gedi} where a language model conditioning on the confounding information $p_{\text{gedi}}(\x|\bc)$ is used to reshape the generation distribution of the above \texttt{Conditional LM}.
We include comparison with more baseline methods in the appendix.

For our approach, the available confounding labels serve as the proxy $\bc$. 
The attribute classifier $f$ used to train our model (Eq.\ref{eq:loss:a}) is pretrained on the biased training data.
%and the reweighted small subset with available confounding labels. 
%, and finetune on the small subset with available confounding labels which allows us to  re-weight the examples according to the correlation strength. 
On \textsc{Yelp}, the resulting (sentiment) classifier has a mediocre accuracy of 83\% on the balanced test set; On \textsc{Bios}, the (gender) classifier has an accuracy of 91\%.

\begin{table*}[t]
%\vspace{-15pt}
\centering
\small
\begin{tabular}{r r c c c c}
\cmidrule[\heavyrulewidth](lr){1-6}
& Methods & {Control accuracy} ($\uparrow$) & {Bias} ($\downarrow$) & {Fluency} ($\downarrow$) & Diversity ($\uparrow$) \\
\cmidrule[\heavyrulewidth](lr){1-6}
\multirow{4}{*}{\textsc{Yelp}} & Conditional LM & 79.1 & 78.7 &  50.4 & 41.4 \\
& Conditional LM (full) & 80.3 & 78.9 & 50.8 &  41.9 \\
& GeDi \citep{krause2020gedi}  & 80.9 & 74.3 & 83.2 &  41.7 \\
\cmidrule(lr){2-6}
& Ablation: {Ours} w/o ${cf\text{-}z/c}$ & 91.1 & 89.2 & 54.1 & 40.4 \\
& {Ours} & {\bf 96.3} & {\bf 59.8} & 51.3 & 39.1 \\
\cmidrule[\heavyrulewidth](lr){1-6} 
\multirow{4}{*}{\textsc{Bios}} & Conditional LM & 95.51 & 84.73 & {\bf 17.0}  & 46.5 \\
& Conditional LM (full) & 93.28 & 72.34 & 18.5 & 48.5 \\
& GeDi \citep{krause2020gedi}  & 86.0 & 75.2 & 27.8 &  43.5 \\
\cmidrule(lr){2-6}
& Ablation: {Ours} w/o ${cf\text{-}z/c}$ & 97.3 & 70.1 & 29.4 & 42.1 \\
& {Ours} & {\bf 99.2} & {\bf 62.4} & 32.0  & 40.6 \\
\cmidrule[\heavyrulewidth](lr){1-6}
\end{tabular}
\vspace{-8pt}
\caption{
Automatic evaluation of attribute-conditional generation on \textsc{Yelp} and \textsc{Bios}. \emph{Control accuracy} is measured by the attribute classifier accuracy; \emph{Bias} is by the confounding classifier accuracy; For \emph{fluency}, we report perplexity, thus a lower score indicates more fluent text; \emph{Diversity} is measured by the Distinct-$2$ metric.
For each evaluation aspect, we highlight the best result that has significant improvements over others.
}
\label{tab:cond-gen-auto}
\vspace{-14pt}
\end{table*}
\begin{table*}[t]
%\vspace{-15pt}
\centering
\small
\begin{tabular}{r r c c c}
\cmidrule[\heavyrulewidth](lr){1-5}
& Methods & {Control accuracy} ($\uparrow$) & {Bias} ($\downarrow$) & {Fluency} ($\uparrow$)  \\
\cmidrule[\heavyrulewidth](lr){1-5}
\multirow{2}{*}{\textsc{Yelp}} & Conditional LM (full) & 80.0  & 73.0  &  3.90 \\
& {Ours} & {\bf 97.0}  &  {\bf 56.0}  & 3.85  \\
\cmidrule[\heavyrulewidth](lr){1-5} 
\multirow{2}{*}{\textsc{Bios}} & Conditional LM (full) & 96.0  & 82.0  & 4.43  \\
& {Ours} & {\bf 99.0} & {\bf 60.0} & 4.25 \\
\cmidrule[\heavyrulewidth](lr){1-5}
\end{tabular}
\vspace{-8pt}
\caption{
Human evaluation of attribute-conditional generation on \textsc{Yelp} and \textsc{Bios}.
}
\label{tab:cond-gen-human}
\vspace{-16pt}
\end{table*}

\paragraph{Evaluation} 
We conduct both automatic and human evaluation. For the former, we follow the common practice and evaluate the generations in terms of various aspects as following: {\bf (1) Control accuracy} for which we use an ``evaluation attribute classifier'' that takes as inputs the generated sentences and measures how accurate they entail the input attributes. The evaluation attribute classifier is trained on a large unbiased set of examples from the original corpus and is of high test accuracy (87\% for \textsc{Yelp} and 95\% on \textsc{Bios}) for evaluation purpose (note the difference from the above classifier $f$ trained with only biased training data); {\bf (2) Bias} which is measured by another classifier for the confounding factor. Intuitively, the better the predicted confounding labels match the input attributes, the more correlated the two factors in the generation. A 50\% match indicates no correlation.
The classifiers are trained similarly as the evaluation attribute classifiers, and achieve accuracy 85\% on \textsc{Yelp} and 90\% for \textsc{Bios}; {\bf (3) Fluency} which is measured by applying GPT-2 language models (LMs) on the generated text and computing the perplexity; The LMs obtain perplexity of 32.4 and 18.0 on the real text of \textsc{Yelp} and \textsc{Bios}, respectively. 
{\bf (4) Diversity} with the common Distinct-$n$ metric \citep{li2016diversity} that measures the ratio of unique $n$-grams against total number of $n$-gram in the generation set. We evaluate 10K generated samples by each model.

For human evaluation, we ask human raters to annotate for each generated text the attribute label and confounding factor label, based on which we compute the control accuracy and bias as above. We also annotate language fluency using a 5-point Likert scale. On each dataset, we compare \texttt{Conditional LM (full)} and our approach, with 100 sentences from each model annotated by 3 raters. The Pearson correlation coefficient of human scores is 0.67, showing strong inter-rater agreement.

\paragraph{Results}

Table~\ref{tab:cond-gen-auto} shows the automatic evaluation results on both \textsc{Yelp} and \textsc{Bios}. Our causal approach significantly improves over the association-based conditional models. For example, on \textsc{Yelp}, our model achieves 16\% absolute improvement in terms of control accuracy, and at the same time reduces the bias (spurious correlation with the confounder) by 19\%. In contrast, the conditional LMs mostly inherit the bias from the training data.
As an ablation study, we also evaluate a simplified variant of our full approach by omitting the counterfactual objectives w.r.t $\z$ and $\bc$ (Eqs.\ref{eq:loss:z} and \ref{eq:loss:c}) (which reduces to a training strategy similar to the previous methods \citep[e.g.,][]{Hu2017TowardCG}). The variant improves the control accuracy over the conditional LMs, but fails to effectively reduce the generation bias. The results show the crucial role of confounder balancing in bias reduction. 
%The results on the \textsc{Bios} dataset  
On the \textsc{Bios} dataset, our approach also obtains consistent improvement on both accuracy and bias.

Table~\ref{tab:cond-gen-human} shows the human evaluation results on both datasets, which largely confirm the above observations with automatic evaluation.

\subsection{Text Attribute Transfer}

We next study text attribute transfer (\S\ref{sec:counterfactual}) as the second core task of controllable generation.
The proposed causal approach also achieves substantial improvement in terms of accurate control and bias reduction. Besides, for a broader comparison, we also apply our approach to another unbiased dataset widely studied in previous text attribute transfer research, showing superior performance.

\paragraph{Datasets} We use the above biased \textsc{Yelp} dataset (\S\ref{sec:exp:attr-gen}) to study the attribute transfer, where we aim to modify a sentence to possess the opposite sentiment (e.g., from negative to positive), and at the same time preserve all other characteristics. In particular, we want the new sentence to keep the category unchanged, which is difficult for previous association-based controllable models given the strong correlation in the data between sentiment and category. Besides, since most previous attribute transfer studies have focused only on unbiased setting, we additionally evaluate our approach on the popular \emph{unbiased} \textsc{Yelp} data (reviews with sentiment for restaurants only) \citep{shen2017style} for comparison. 

\paragraph{Evaluation}
We follow the standard practice for evaluation. For the biased setting, we measure {\bf control accuracy}, {\bf bias}, and {\bf fluency} as in \S\ref{sec:exp:attr-gen}. We also assess the common aspect {\bf preservation}, which evaluates the BLEU score \citep{papineni2002bleu} between the generated and original sentences (i.e., self-BLEU). A higher score indicates better preservation of sentence properties. For the unbiased setting, we omit the bias evaluation, and additionally compute another preservation metric, ref-BLEU, which is the BLEU score between the generation and human-written golden text on a subset of test examples \citep{li2018delete}. We also conduct {\bf human evaluation} which shows the same conclusions as the automatic evaluation in terms of model performance. We put the results in appendix due to space limitation. 

%We compare with existing text attribute transfer methods, which 

\paragraph{Results}
Table~\ref{tab:cond-gen-tst} shows results on the biased \textsc{Yelp} data, a substantially more challenging setting than the popular unbiased one (Table~\ref{tab:cond-gen-tst-unbias}). We compare two of the previous best-performing methods with public code. Our approach again manages to reduce the bias while achieving decent transfer accuracy. The previous methods struggle to edit the text on many instances (e.g., generating the same sentences as inputs), leading to low control accuracy. Ablation comparison with our simplified variant (\texttt{our w/o cf-z/c}) further validates the effect of counterfactual objectives for confounder balancing (\S\ref{sec:learning}), as shown by the improved accuracy and mitigated bias of the full approach. 

Finally, Table~\ref{tab:cond-gen-tst-unbias} shows the results on the common unbiased \textsc{Yelp} sentiment data.
%, serving as a reference as most previous methods report performance on the benchmark. 
The results show our approach generates fluent output with improved accuracy and preservation.

\begin{table*}[t]
%\vspace{-15pt}
\centering
\small
\begin{tabular}{r c c c c}
\cmidrule[\heavyrulewidth](lr){1-5}
Methods & {Control accuracy} ($\uparrow$) & {Bias} ($\downarrow$) & Preservation ($\uparrow$) & {Fluency} ($\downarrow$) \\
\cmidrule[\heavyrulewidth](lr){1-5}
\citet{Hu2017TowardCG} & 44.1  &  68.4  & 77.7 & 132.7 \\
\citet{he2019probabilistic} & 35.3 & 60.2 & {\bf 80.1} & 57.7 \\
\cmidrule(lr){1-5}
Ablation: {Ours} w/o ${cf\text{-}z/c}$ & 75.0 & 67.8  & 36.3 & 34.2 \\
{Ours} & {\bf 77.0}  & 61.4 & 42.3 & {\bf 29.6} \\
\cmidrule[\heavyrulewidth](lr){1-5} 
\end{tabular}
\vspace{-8pt}
\caption{
Results of attribute transfer on the \emph{biased} \textsc{Yelp}. Baselines \citep{Hu2017TowardCG,he2019probabilistic} with the public code, and they fail to rewrite the text on most instances, leading to very low control accuracy and high preservation.
}
\label{tab:cond-gen-tst}
\vspace{-12pt}
\end{table*}
\begin{table*}[t]
%\vspace{-15pt}
\centering
\small
\begin{tabular}{r c c c c}
\cmidrule[\heavyrulewidth](lr){1-5}
 \multirow{2}{*}{Methods} & \multirow{2}{*}{Control accuracy ($\uparrow$)}  & \multicolumn{2}{c}{Preservation ($\uparrow$)} & \multirow{2}{*}{Fluency ($\downarrow$)}  \\
&  &  {\scriptsize self-BLEU} & {\scriptsize ref-BLEU} \\
\cmidrule[\heavyrulewidth](lr){1-5}
\citet{Hu2017TowardCG} & 86.7 & {\bf 58.4} & - & 177.7  \\
\citet{shen2017style} & 73.9 & 20.7 & 7.8 & 72.0   \\ 
 \citet{he2019probabilistic} & 87.9 &  48.4 & 18.7 & {\bf 31.7} \\
 \citet{dai2019style} & 87.7 & 54.9 & 20.3 & 73.0 \\
\cmidrule(lr){1-5}
 Ablation: {Ours} w/o ${cf\text{-}z/c}$ & 87.1  & 57.2  & 24.3 & 46.6 \\
%{Ours} & 90.1  &  59.0 & 24.8 & -41.1 \\
{Ours} & {\bf 91.9}  &  57.3 & {\bf 25.5} & 47.1 \\
\cmidrule[\heavyrulewidth](lr){1-5} 
\end{tabular}
\vspace{-6pt}
\caption{
Results of text attribute transfer on the common \emph{unbiased} \textsc{Yelp} data.
}
\label{tab:cond-gen-tst-unbias}
\vspace{-15pt}
\end{table*}

% \subsection{Debiasing Data and Pretrained Language Models}

% Finally 

\section{Conclusions and Future Work}

We have presented a principled causal perspective for the two core tasks of controllable text generation. Based on the proposed structural causal model, attribute-conditional generation is modeled as interventional inference, and text attribute transfer performs counterfactual prediction. We connect rich techniques in causality, disentangled representation, and text generative modeling, and develop learning objectives for accurate control and confounder balancing. Focusing on the challenging setting with partially available confounding information, the experiments show our approach achieves accurate control and mitigates the strong correlations in the data. 

The proposed causal framework opens up a range of new opportunities for further improving and enriching controllable text generation. For example, though this work has focused on  single control attribute and confounding factor, it would be interesting to generalize the approach for structured control of a richer set of text attributes, by modeling the underlying {\bf causal graph between attributes} (as explored similarly in image generation \citep{yang2020causalvae,shen2020disentangled}). Besides, we are interested in importing more causality tools through the causal perspective to enable new applications. For instance, the \emph{inverse propensity reweighting} technique in causality can potentially be used to {\bf debias pretrained language models} $p_{\text{pretrain}}(\x|a)$, with the following known equation between the unbiased interventional conditional $p(\x|do(a))$ and the biased standard conditional $p(\x|a)$:
\begin{equation}
\small
\begin{split}
    p(\x | do(a)) 
    &= \sum\nolimits_{\z} p(\x | a, \z) p(\z) 
    %&= \sum\nolimits_{\z} p(\x | a, \z) p(\z|a)\frac{p(a)}{p(a|\z)}  \\
    = \sum\nolimits_{\z} p(\x|a) p(\z|\x, a) \frac{p(a)}{p(a|\z)}, %= \sum\nolimits_{\z} p_{\lm}(\x|a) q_\phi(\z|\x, a, \bc) \frac{p(a)}{p_\theta(a|\z)},
\end{split}
\label{eq:propensity}
\end{equation}
where $p(a|\z)$ is known as the \emph{propensity score} \citep{pearl2009causality}, i.e., the propensity (probability) of the $\z$ being assigned to the particular treatment $a$. Plugging in the $p_{\text{pretrain}}(\x|a)$ together with the parameterized estimates of $p_\theta(\z|\x,a)$ and $p_\theta(a|\z)$ as learned in \S\ref{sec:method}, we would effectively convert the pretrained LM into the unbiased $p(\x | do(a))$. Further, rich studies in the causality literature have proposed stabilized and enhanced variants of the above inverse propensity reweighting \citep[e.g., see][]{yao2020survey}, all of which present interesting topics to explore in the controllable generation setting in the future.

\paragraph{Ethical considerations}

We would like to note that automatic text generation could be used maliciously to generate fake, toxic, or offensive content \citep{kreps2020all,wallace2019universal,bender2021dangers}. We hope the unbiased modeling study could offer techniques to alleviate potential issues.

\bibliographystyle{abbrvnat}
\bibliography{custom}

\begin{thebibliography}{86}
\providecommand{\natexlab}[1]{#1}
\providecommand{\url}[1]{\texttt{#1}}
\expandafter\ifx\csname urlstyle\endcsname\relax
  \providecommand{\doi}[1]{doi: #1}\else
  \providecommand{\doi}{doi: \begingroup \urlstyle{rm}\Url}\fi

\bibitem[Angrist and Pischke(2008)]{angrist2008mostly}
J.~D. Angrist and J.-S. Pischke.
\newblock \emph{Mostly harmless econometrics: An empiricist's companion}.
\newblock Princeton university press, 2008.

\bibitem[Bender et~al.(2021)Bender, Gebru, McMillan-Major, and
  Shmitchell]{bender2021dangers}
E.~M. Bender, T.~Gebru, A.~McMillan-Major, and S.~Shmitchell.
\newblock On the dangers of stochastic parrots: Can language models be too big?
\newblock In \emph{Proceedings of the 2021 ACM Conference on Fairness,
  Accountability, and Transparency}, pages 610--623, 2021.

\bibitem[Besserve et~al.(2020)Besserve, Mehrjou, Sun, and
  Sch{\"o}lkopf]{besserve2020counterfactuals}
M.~Besserve, A.~Mehrjou, R.~Sun, and B.~Sch{\"o}lkopf.
\newblock Counterfactuals uncover the modular structure of deep generative
  models.
\newblock In \emph{Eighth International Conference on Learning Representations
  (ICLR 2020)}, 2020.

\bibitem[Bowman et~al.(2016)Bowman, Vilnis, Vinyals, Dai, Jozefowicz, and
  Bengio]{bowman-etal-2016-generating}
S.~R. Bowman, L.~Vilnis, O.~Vinyals, A.~Dai, R.~Jozefowicz, and S.~Bengio.
\newblock Generating sentences from a continuous space.
\newblock In \emph{Proceedings of The 20th {SIGNLL} Conference on Computational
  Natural Language Learning}, pages 10--21, Berlin, Germany, Aug. 2016.
  Association for Computational Linguistics.
\newblock \doi{10.18653/v1/K16-1002}.
\newblock URL \url{https://www.aclweb.org/anthology/K16-1002}.

\bibitem[Brown et~al.(2020)Brown, Mann, Ryder, Subbiah, Kaplan, Dhariwal,
  Neelakantan, Shyam, Sastry, Askell, et~al.]{brown2020language}
T.~B. Brown, B.~Mann, N.~Ryder, M.~Subbiah, J.~Kaplan, P.~Dhariwal,
  A.~Neelakantan, P.~Shyam, G.~Sastry, A.~Askell, et~al.
\newblock Language models are few-shot learners.
\newblock \emph{arXiv preprint arXiv:2005.14165}, 2020.

\bibitem[Chattopadhyay et~al.(2019)Chattopadhyay, Manupriya, Sarkar, and
  Balasubramanian]{chattopadhyay2019neural}
A.~Chattopadhyay, P.~Manupriya, A.~Sarkar, and V.~N. Balasubramanian.
\newblock Neural network attributions: A causal perspective.
\newblock In \emph{International Conference on Machine Learning}, pages
  981--990. PMLR, 2019.

\bibitem[Chen et~al.(2020)Chen, Tian, Xiao, He, and
  Jin]{chen-etal-2020-exploring}
W.~Chen, J.~Tian, L.~Xiao, H.~He, and Y.~Jin.
\newblock Exploring logically dependent multi-task learning with causal
  inference.
\newblock In \emph{Proceedings of the 2020 Conference on Empirical Methods in
  Natural Language Processing (EMNLP)}, 2020.

\bibitem[Dai et~al.(2019)Dai, Liang, Qiu, and Huang]{dai2019style}
N.~Dai, J.~Liang, X.~Qiu, and X.-J. Huang.
\newblock Style transformer: Unpaired text style transfer without disentangled
  latent representation.
\newblock In \emph{Proceedings of the 57th Annual Meeting of the Association
  for Computational Linguistics}, pages 5997--6007, 2019.

\bibitem[Dathathri et~al.(2020)Dathathri, Madotto, Lan, Hung, Frank, Molino,
  Yosinski, and Liu]{dathathri2020plug}
S.~Dathathri, A.~Madotto, J.~Lan, J.~Hung, E.~Frank, P.~Molino, J.~Yosinski,
  and R.~Liu.
\newblock Plug and play language models: {A} simple approach to controlled text
  generation.
\newblock In \emph{8th International Conference on Learning Representations,
  {ICLR} 2020, Addis Ababa, Ethiopia, April 26-30, 2020}. OpenReview.net, 2020.
\newblock URL \url{https://openreview.net/forum?id=H1edEyBKDS}.

\bibitem[De-Arteaga et~al.(2019)De-Arteaga, Romanov, Wallach, Chayes, Borgs,
  Chouldechova, Geyik, Kenthapadi, and Kalai]{de2019bias}
M.~De-Arteaga, A.~Romanov, H.~Wallach, J.~Chayes, C.~Borgs, A.~Chouldechova,
  S.~Geyik, K.~Kenthapadi, and A.~T. Kalai.
\newblock Bias in bios: A case study of semantic representation bias in a
  high-stakes setting.
\newblock In \emph{proceedings of the Conference on Fairness, Accountability,
  and Transparency}, pages 120--128, 2019.

\bibitem[Dinan et~al.(2020)Dinan, Fan, Williams, Urbanek, Kiela, and
  Weston]{dinan2020queens}
E.~Dinan, A.~Fan, A.~Williams, J.~Urbanek, D.~Kiela, and J.~Weston.
\newblock Queens are powerful too: Mitigating gender bias in dialogue
  generation.
\newblock In \emph{Proceedings of the 2020 Conference on Empirical Methods in
  Natural Language Processing (EMNLP)}, pages 8173--8188, 2020.

\bibitem[Egami et~al.(2018)Egami, Fong, Grimmer, Roberts, and
  Stewart]{egami2018make}
N.~Egami, C.~J. Fong, J.~Grimmer, M.~E. Roberts, and B.~M. Stewart.
\newblock How to make causal inferences using texts.
\newblock \emph{arXiv preprint arXiv:1802.02163}, 2018.

\bibitem[Feder et~al.(2020)Feder, Oved, Shalit, and Reichart]{feder2020causalm}
A.~Feder, N.~Oved, U.~Shalit, and R.~Reichart.
\newblock Causa{LM}: Causal model explanation through counterfactual language
  models.
\newblock \emph{arXiv preprint arXiv:2005.13407}, 2020.

\bibitem[Ficler and Goldberg(2017)]{ficler2017controlling}
J.~Ficler and Y.~Goldberg.
\newblock Controlling linguistic style aspects in neural language generation.
\newblock \emph{CoRR}, abs/1707.02633, 2017.
\newblock URL \url{http://arxiv.org/abs/1707.02633}.

\bibitem[Gan et~al.(2021)Gan, Li, Lipton, and Tayur]{gan2021causal}
K.~Gan, A.~Li, Z.~Lipton, and S.~Tayur.
\newblock Causal inference with selectively deconfounded data.
\newblock In \emph{International Conference on Artificial Intelligence and
  Statistics}, pages 2791--2799. PMLR, 2021.

\bibitem[Goodfellow et~al.(2014)Goodfellow, Pouget-Abadie, Mirza, Xu,
  Warde-Farley, Ozair, Courville, and Bengio]{Goodfellow2014GenerativeAN}
I.~J. Goodfellow, J.~Pouget-Abadie, M.~Mirza, B.~Xu, D.~Warde-Farley, S.~Ozair,
  A.~C. Courville, and Y.~Bengio.
\newblock Generative adversarial nets.
\newblock In \emph{NIPS}, 2014.

\bibitem[Goyal et~al.(2019)Goyal, Feder, Shalit, and Kim]{goyal2019explaining}
Y.~Goyal, A.~Feder, U.~Shalit, and B.~Kim.
\newblock Explaining classifiers with causal concept effect (cace).
\newblock \emph{arXiv preprint arXiv:1907.07165}, 2019.

\bibitem[Guo et~al.(2021)Guo, Tan, Liu, Xing, and Hu]{guo2021text}
H.~Guo, B.~Tan, Z.~Liu, E.~P. Xing, and Z.~Hu.
\newblock Text generation with efficient (soft) {Q}-learning.
\newblock \emph{arXiv preprint arXiv:2106.07704}, 2021.

\bibitem[Hassanpour and Greiner(2020)]{hassanpour2020learning}
N.~Hassanpour and R.~Greiner.
\newblock Learning disentangled representations for counterfactual regression.
\newblock In \emph{International Conference on Learning Representations}, 2020.

\bibitem[He et~al.(2019{\natexlab{a}})He, Wang, Neubig, and
  Berg-Kirkpatrick]{he2019probabilistic}
J.~He, X.~Wang, G.~Neubig, and T.~Berg-Kirkpatrick.
\newblock A probabilistic formulation of unsupervised text style transfer.
\newblock In \emph{International Conference on Learning Representations
  (ICLR)}, 2019{\natexlab{a}}.

\bibitem[He et~al.(2019{\natexlab{b}})He, Zuo, Kan, Shan, and
  Chen]{he2019attgan}
Z.~He, W.~Zuo, M.~Kan, S.~Shan, and X.~Chen.
\newblock Attgan: Facial attribute editing by only changing what you want.
\newblock \emph{IEEE Transactions on Image Processing}, 28\penalty0
  (11):\penalty0 5464--5478, 2019{\natexlab{b}}.

\bibitem[Hu et~al.(2017)Hu, Yang, Liang, Salakhutdinov, and
  Xing]{Hu2017TowardCG}
Z.~Hu, Z.~Yang, X.~Liang, R.~Salakhutdinov, and E.~Xing.
\newblock Toward controlled generation of text.
\newblock In \emph{International Conference on Machine Learning (ICML)}, 2017.

\bibitem[Huang et~al.(2020)Huang, Zhang, Jiang, Stanforth, Welbl, Rae, Maini,
  Yogatama, and Kohli]{huang2020reducing}
P.-S. Huang, H.~Zhang, R.~Jiang, R.~Stanforth, J.~Welbl, J.~Rae, V.~Maini,
  D.~Yogatama, and P.~Kohli.
\newblock Reducing sentiment bias in language models via counterfactual
  evaluation.
\newblock In \emph{Proceedings of the 2020 Conference on Empirical Methods in
  Natural Language Processing: Findings}, pages 65--83, 2020.

\bibitem[Ilse et~al.(2021)Ilse, Tomczak, Forr{\'e}, et~al.]{ilse2020selecting}
M.~Ilse, J.~Tomczak, P.~Forr{\'e}, et~al.
\newblock Selecting data augmentation for simulating interventions.
\newblock In \emph{AAAI}, 2021.

\bibitem[Jang et~al.(2016)Jang, Gu, and Poole]{jang2016categorical}
E.~Jang, S.~Gu, and B.~Poole.
\newblock Categorical reparameterization with gumbel-softmax.
\newblock \emph{arXiv preprint arXiv:1611.01144}, 2016.

\bibitem[Jin et~al.(2020)Jin, Jin, Hu, Vechtomova, and Mihalcea]{jin2020deep}
D.~Jin, Z.~Jin, Z.~Hu, O.~Vechtomova, and R.~Mihalcea.
\newblock Deep learning for text style transfer: A survey.
\newblock \emph{arXiv preprint arXiv:2011.00416}, 2020.

\bibitem[Johansson et~al.(2016)Johansson, Shalit, and
  Sontag]{johansson2016learning}
F.~Johansson, U.~Shalit, and D.~Sontag.
\newblock Learning representations for counterfactual inference.
\newblock In \emph{International conference on machine learning}, pages
  3020--3029. PMLR, 2016.

\bibitem[Kaushik et~al.(2019)Kaushik, Hovy, and Lipton]{kaushik2019learning}
D.~Kaushik, E.~Hovy, and Z.~Lipton.
\newblock Learning the difference that makes a difference with
  counterfactually-augmented data.
\newblock In \emph{International Conference on Learning Representations}, 2019.

\bibitem[Keith et~al.(2020)Keith, Jensen, and O’Connor]{keith2020text}
K.~Keith, D.~Jensen, and B.~O’Connor.
\newblock Text and causal inference: A review of using text to remove
  confounding from causal estimates.
\newblock In \emph{Proceedings of the 58th Annual Meeting of the Association
  for Computational Linguistics}, pages 5332--5344, 2020.

\bibitem[Keskar et~al.(2019)Keskar, McCann, Varshney, Xiong, and
  Socher]{keskar2019ctrl}
N.~S. Keskar, B.~McCann, L.~R. Varshney, C.~Xiong, and R.~Socher.
\newblock Ctrl: A conditional transformer language model for controllable
  generation.
\newblock \emph{arXiv preprint arXiv:1909.05858}, 2019.

\bibitem[Kingma and Welling(2013)]{kingma2013auto}
D.~P. Kingma and M.~Welling.
\newblock Auto-encoding variational bayes.
\newblock \emph{arXiv preprint arXiv:1312.6114}, 2013.

\bibitem[Kocaoglu et~al.(2018)Kocaoglu, Snyder, Dimakis, and
  Vishwanath]{kocaoglu2018causalgan}
M.~Kocaoglu, C.~Snyder, A.~G. Dimakis, and S.~Vishwanath.
\newblock Causal{GAN}: Learning causal implicit generative models with
  adversarial training.
\newblock In \emph{International Conference on Learning Representations}, 2018.

\bibitem[Krause et~al.(2020)Krause, Gotmare, McCann, Keskar, Joty, Socher, and
  Rajani]{krause2020gedi}
B.~Krause, A.~D. Gotmare, B.~McCann, N.~S. Keskar, S.~Joty, R.~Socher, and
  N.~F. Rajani.
\newblock Gedi: Generative discriminator guided sequence generation.
\newblock \emph{arXiv preprint arXiv:2009.06367}, 2020.

\bibitem[Kreps et~al.(2020)Kreps, McCain, and Brundage]{kreps2020all}
S.~Kreps, R.~M. McCain, and M.~Brundage.
\newblock All the news that’s fit to fabricate: Ai-generated text as a tool
  of media misinformation.
\newblock \emph{Journal of Experimental Political Science}, pages 1--14, 2020.

\bibitem[Lample et~al.(2017)Lample, Zeghidour, Usunier, Bordes, Denoyer, and
  Ranzato]{lample2017fader}
G.~Lample, N.~Zeghidour, N.~Usunier, A.~Bordes, L.~Denoyer, and M.~Ranzato.
\newblock Fader networks: Manipulating images by sliding attributes.
\newblock In I.~Guyon, U.~von Luxburg, S.~Bengio, H.~M. Wallach, R.~Fergus,
  S.~V.~N. Vishwanathan, and R.~Garnett, editors, \emph{Advances in Neural
  Information Processing Systems 30: Annual Conference on Neural Information
  Processing Systems 2017, 4-9 December 2017, Long Beach, CA, {USA}}, pages
  5967--5976, 2017.
\newblock URL
  \url{http://papers.nips.cc/paper/7178-fader-networksmanipulating-images-by-sliding-attributes}.

\bibitem[Li et~al.(2020)Li, Gao, Li, Li, Peng, Zhang, and Gao]{li2020optimus}
C.~Li, X.~Gao, Y.~Li, X.~Li, B.~Peng, Y.~Zhang, and J.~Gao.
\newblock Optimus: Organizing sentences via pre-trained modeling of a latent
  space.
\newblock \emph{arXiv preprint arXiv:2004.04092}, 2020.

\bibitem[Li et~al.(2016)Li, Galley, Brockett, Gao, and Dolan]{li2016diversity}
J.~Li, M.~Galley, C.~Brockett, J.~Gao, and W.~B. Dolan.
\newblock A diversity-promoting objective function for neural conversation
  models.
\newblock In \emph{Proceedings of the 2016 Conference of the North American
  Chapter of the Association for Computational Linguistics: Human Language
  Technologies}, pages 110--119, 2016.

\bibitem[Li et~al.(2018)Li, Jia, He, and Liang]{li2018delete}
J.~Li, R.~Jia, H.~He, and P.~Liang.
\newblock Delete, retrieve, generate: A simple approach to sentiment and style
  transfer.
\newblock In \emph{2018 Conference of the North American Chapter of the
  Association for Computational Linguistics: Human Language Technologies, NAACL
  HLT 2018}, pages 1865--1874. Association for Computational Linguistics (ACL),
  2018.

\bibitem[Li et~al.(2021)Li, Yavuz, Hashimoto, Li, Niu, Rajani, Yan, Zhou, and
  Xiong]{li2020coco}
S.~Li, S.~Yavuz, K.~Hashimoto, J.~Li, T.~Niu, N.~Rajani, X.~Yan, Y.~Zhou, and
  C.~Xiong.
\newblock Coco: Controllable counterfactuals for evaluating dialogue state
  trackers.
\newblock In \emph{ICLR}, 2021.

\bibitem[Liu et~al.(2021)Liu, Jia, Wei, Xu, Wang, and
  Vosoughi]{liu2021mitigating}
R.~Liu, C.~Jia, J.~Wei, G.~Xu, L.~Wang, and S.~Vosoughi.
\newblock Mitigating political bias in language models through reinforced
  calibration.
\newblock In \emph{Proceedings of the AAAI Conference on Artificial
  Intelligence}, 2021.

\bibitem[Locatello et~al.(2019)Locatello, Tschannen, Bauer, R{\"a}tsch,
  Sch{\"o}lkopf, and Bachem]{locatello2019disentangling}
F.~Locatello, M.~Tschannen, S.~Bauer, G.~R{\"a}tsch, B.~Sch{\"o}lkopf, and
  O.~Bachem.
\newblock Disentangling factors of variations using few labels.
\newblock In \emph{International Conference on Learning Representations}, 2019.

\bibitem[Loshchilov and Hutter(2018)]{loshchilov2018decoupled}
I.~Loshchilov and F.~Hutter.
\newblock Decoupled weight decay regularization.
\newblock In \emph{International Conference on Learning Representations}, 2018.

\bibitem[Louizos et~al.(2017)Louizos, Shalit, Mooij, Sontag, Zemel, and
  Welling]{louizos2017causal}
C.~Louizos, U.~Shalit, J.~M. Mooij, D.~Sontag, R.~Zemel, and M.~Welling.
\newblock Causal effect inference with deep latent-variable models.
\newblock In \emph{Advances in neural information processing systems
  (NeurIPS)}, pages 6446--6456, 2017.

\bibitem[Lu et~al.(2020)Lu, Tao, Chen, Li, Guo, and Carin]{lu2018considering}
D.~Lu, C.~Tao, J.~Chen, F.~Li, F.~Guo, and L.~Carin.
\newblock Reconsidering generative objectives for counterfactual reasoning.
\newblock In \emph{NeurIPS}, 2020.

\bibitem[Luo et~al.(2019)Luo, Li, Zhou, Yang, Chang, Sui, and Sun]{luo2019dual}
F.~Luo, P.~Li, J.~Zhou, P.~Yang, B.~Chang, Z.~Sui, and X.~Sun.
\newblock A dual reinforcement learning framework for unsupervised text style
  transfer.
\newblock In \emph{IJCAI}, 2019.

\bibitem[Madaan et~al.(2021)Madaan, Padhi, Panwar, and
  Saha]{madaan2021generate}
N.~Madaan, I.~Padhi, N.~Panwar, and D.~Saha.
\newblock Generate your counterfactuals: Towards controlled counterfactual
  generation for text.
\newblock In \emph{AAAI}, 2021.

\bibitem[Maddison et~al.(2017)Maddison, Mnih, and Teh]{maddison2016concrete}
C.~J. Maddison, A.~Mnih, and Y.~W. Teh.
\newblock The concrete distribution: A continuous relaxation of discrete random
  variables.
\newblock In \emph{ICLR}, 2017.

\bibitem[Madras et~al.(2019)Madras, Creager, Pitassi, and
  Zemel]{madras2019fairness}
D.~Madras, E.~Creager, T.~Pitassi, and R.~Zemel.
\newblock Fairness through causal awareness: Learning causal latent-variable
  models for biased data.
\newblock In \emph{Proceedings of the Conference on Fairness, Accountability,
  and Transparency}, pages 349--358, 2019.

\bibitem[Mani and Cooper(2000)]{mani2000causal}
S.~Mani and G.~F. Cooper.
\newblock Causal discovery from medical textual data.
\newblock In \emph{Proceedings of the AMIA Symposium}, page 542. American
  Medical Informatics Association, 2000.

\bibitem[Montgomery et~al.(2000)Montgomery, Gragnolati, Burke, and
  Paredes]{montgomery2000measuring}
M.~R. Montgomery, M.~Gragnolati, K.~A. Burke, and E.~Paredes.
\newblock Measuring living standards with proxy variables.
\newblock \emph{Demography}, 37\penalty0 (2):\penalty0 155--174, 2000.

\bibitem[Moraffah et~al.(2020)Moraffah, Moraffah, Karami, Raglin, and
  Liu]{moraffah2020can}
R.~Moraffah, B.~Moraffah, M.~Karami, A.~Raglin, and H.~Liu.
\newblock Can: A causal adversarial network for learning observational and
  interventional distributions.
\newblock \emph{arXiv preprint arXiv:2008.11376}, 2020.

\bibitem[Ozery-Flato et~al.(2018)Ozery-Flato, Thodoroff, Ninio, Rosen-Zvi, and
  El-Hay]{ozery2018adversarial}
M.~Ozery-Flato, P.~Thodoroff, M.~Ninio, M.~Rosen-Zvi, and T.~El-Hay.
\newblock Adversarial balancing for causal inference.
\newblock \emph{arXiv preprint arXiv:1810.07406}, 2018.

\bibitem[Papineni et~al.(2002)Papineni, Roukos, Ward, and
  Zhu]{papineni2002bleu}
K.~Papineni, S.~Roukos, T.~Ward, and W.-J. Zhu.
\newblock Bleu: a method for automatic evaluation of machine translation.
\newblock In \emph{Proceedings of the 40th annual meeting of the Association
  for Computational Linguistics}, pages 311--318, 2002.

\bibitem[Parascandolo et~al.(2018)Parascandolo, Kilbertus, Rojas-Carulla, and
  Sch{\"o}lkopf]{parascandolo2018learning}
G.~Parascandolo, N.~Kilbertus, M.~Rojas-Carulla, and B.~Sch{\"o}lkopf.
\newblock Learning independent causal mechanisms.
\newblock In \emph{International Conference on Machine Learning}, pages
  4036--4044. PMLR, 2018.

\bibitem[Pawlowski et~al.(2020)Pawlowski, Coelho~de Castro, and
  Glocker]{pawlowski2020deep}
N.~Pawlowski, D.~Coelho~de Castro, and B.~Glocker.
\newblock Deep structural causal models for tractable counterfactual inference.
\newblock \emph{Advances in Neural Information Processing Systems}, 33, 2020.

\bibitem[Pearl(2009)]{pearl2009causality}
J.~Pearl.
\newblock \emph{Causality}.
\newblock Cambridge university press, 2009.

\bibitem[Prates et~al.(2019)Prates, Avelar, and Lamb]{prates2019assessing}
M.~O. Prates, P.~H. Avelar, and L.~C. Lamb.
\newblock Assessing gender bias in machine translation: a case study with
  google translate.
\newblock \emph{Neural Computing and Applications}, pages 1--19, 2019.

\bibitem[Qin et~al.(2019)Qin, Bosselut, Holtzman, Bhagavatula, Clark, and
  Choi]{qin2019counterfactual}
L.~Qin, A.~Bosselut, A.~Holtzman, C.~Bhagavatula, E.~Clark, and Y.~Choi.
\newblock Counterfactual story reasoning and generation.
\newblock In \emph{EMNLP}, pages 5046--5056, 2019.

\bibitem[Radford et~al.(2019)Radford, Wu, Child, Luan, Amodei, and
  Sutskever]{radford2019language}
A.~Radford, J.~Wu, R.~Child, D.~Luan, D.~Amodei, and I.~Sutskever.
\newblock Language models are unsupervised multitask learners.
\newblock \emph{OpenAI Blog}, 1\penalty0 (8):\penalty0 9, 2019.

\bibitem[Reiter and Dale(1997)]{reiter1997building}
E.~Reiter and R.~Dale.
\newblock Building applied natural language generation systems.
\newblock \emph{Natural Language Engineering}, 3\penalty0 (1):\penalty0 57--87,
  1997.

\bibitem[Sauer and Geiger(2021)]{sauer2021counterfactual}
A.~Sauer and A.~Geiger.
\newblock Counterfactual generative networks.
\newblock In \emph{ICLR}, 2021.

\bibitem[Sch{\"o}lkopf(2019)]{scholkopf2019causality}
B.~Sch{\"o}lkopf.
\newblock Causality for machine learning.
\newblock \emph{arXiv preprint arXiv:1911.10500}, 2019.

\bibitem[Shalit et~al.(2017)Shalit, Johansson, and
  Sontag]{shalit2017estimating}
U.~Shalit, F.~D. Johansson, and D.~Sontag.
\newblock Estimating individual treatment effect: generalization bounds and
  algorithms.
\newblock In \emph{International Conference on Machine Learning}, pages
  3076--3085. PMLR, 2017.

\bibitem[Shen et~al.(2017)Shen, Lei, Barzilay, and Jaakkola]{shen2017style}
T.~Shen, T.~Lei, R.~Barzilay, and T.~Jaakkola.
\newblock Style transfer from non-parallel text by cross-alignment.
\newblock In \emph{Advances in neural information processing systems
  (NeurIPS)}, pages 6830--6841, 2017.

\bibitem[Shen et~al.(2020)Shen, Liu, Dong, Lian, Chen, and
  Zhang]{shen2020disentangled}
X.~Shen, F.~Liu, H.~Dong, Q.~Lian, Z.~Chen, and T.~Zhang.
\newblock Disentangled generative causal representation learning.
\newblock \emph{arXiv preprint arXiv:2010.02637}, 2020.

\bibitem[Sheng et~al.(2019)Sheng, Chang, Natarajan, and Peng]{sheng2019woman}
E.~Sheng, K.-W. Chang, P.~Natarajan, and N.~Peng.
\newblock The woman worked as a babysitter: On biases in language generation.
\newblock In \emph{EMNLP}, pages 3398--3403, 2019.

\bibitem[Sheng et~al.(2020)Sheng, Chang, Natarajan, and Peng]{sheng2020towards}
E.~Sheng, K.-W. Chang, P.~Natarajan, and N.~Peng.
\newblock Towards controllable biases in language generation.
\newblock In \emph{Proceedings of the 2020 Conference on Empirical Methods in
  Natural Language Processing: Findings}, pages 3239--3254, 2020.

\bibitem[Stafanovi{\v{c}}s et~al.(2020)Stafanovi{\v{c}}s, Pinnis, and
  Bergmanis]{stafanovivcs2020mitigating}
A.~Stafanovi{\v{c}}s, M.~Pinnis, and T.~Bergmanis.
\newblock Mitigating gender bias in machine translation with target gender
  annotations.
\newblock In \emph{Proceedings of the Fifth Conference on Machine Translation},
  pages 629--638, 2020.

\bibitem[Stanovsky et~al.(2019)Stanovsky, Smith, and
  Zettlemoyer]{stanovsky2019evaluating}
G.~Stanovsky, N.~A. Smith, and L.~Zettlemoyer.
\newblock Evaluating gender bias in machine translation.
\newblock In \emph{Proceedings of the 57th Annual Meeting of the Association
  for Computational Linguistics}, pages 1679--1684, 2019.

\bibitem[Stock et~al.(2012)Stock, Watson, et~al.]{stock2012introduction}
J.~H. Stock, M.~W. Watson, et~al.
\newblock \emph{Introduction to econometrics}, volume~3.
\newblock Pearson New York, 2012.

\bibitem[Tan et~al.(2014)Tan, Lee, and Pang]{tan2014effect}
C.~Tan, L.~Lee, and B.~Pang.
\newblock The effect of wording on message propagation: Topic-and
  author-controlled natural experiments on twitter.
\newblock In \emph{Proceedings of the 52nd Annual Meeting of the Association
  for Computational Linguistics (Volume 1: Long Papers)}, pages 175--185, 2014.

\bibitem[Wallace et~al.(2019)Wallace, Feng, Kandpal, Gardner, and
  Singh]{wallace2019universal}
E.~Wallace, S.~Feng, N.~Kandpal, M.~Gardner, and S.~Singh.
\newblock Universal adversarial triggers for attacking and analyzing nlp.
\newblock In \emph{EMNLP}, 2019.

\bibitem[Wang et~al.(2020)Wang, Qinami, Karakozis, Genova, Nair, Hata, and
  Russakovsky]{wang2020towards}
Z.~Wang, K.~Qinami, I.~C. Karakozis, K.~Genova, P.~Nair, K.~Hata, and
  O.~Russakovsky.
\newblock Towards fairness in visual recognition: Effective strategies for bias
  mitigation.
\newblock In \emph{Proceedings of the IEEE/CVF Conference on Computer Vision
  and Pattern Recognition}, pages 8919--8928, 2020.

\bibitem[Weber et~al.(2020)Weber, Rudinger, and Van~Durme]{weber2020causal}
N.~Weber, R.~Rudinger, and B.~Van~Durme.
\newblock Causal inference of script knowledge.
\newblock In \emph{Proceedings of the 2020 Conference on Empirical Methods in
  Natural Language Processing (EMNLP)}, pages 7583--7596, 2020.

\bibitem[Wood-Doughty et~al.(2018)Wood-Doughty, Shpitser, and
  Dredze]{wood2018challenges}
Z.~Wood-Doughty, I.~Shpitser, and M.~Dredze.
\newblock Challenges of using text classifiers for causal inference.
\newblock In \emph{EMNLP}, volume 2018, page 4586. NIH Public Access, 2018.

\bibitem[Wu et~al.()Wu, Ribeiro, Heer, and Weld]{wu2021polyjuice}
T.~Wu, M.~T. Ribeiro, J.~Heer, and D.~S. Weld.
\newblock Polyjuice: Automated, general-purpose counterfactual generation.
\newblock \emph{arXiv preprint arXiv:2101.00288}.

\bibitem[Wu et~al.(2020)Wu, Kuang, Zhang, Liu, Sun, Xiao, Zhuang, Si, and
  Wu]{wu2020biased}
Y.~Wu, K.~Kuang, Y.~Zhang, X.~Liu, C.~Sun, J.~Xiao, Y.~Zhuang, L.~Si, and
  F.~Wu.
\newblock De-biased court’s view generation with causality.
\newblock In \emph{Proceedings of the 2020 Conference on Empirical Methods in
  Natural Language Processing (EMNLP)}, pages 763--780, 2020.

\bibitem[Yang et~al.(2020)Yang, Liu, Chen, Shen, Hao, and
  Wang]{yang2020causalvae}
M.~Yang, F.~Liu, Z.~Chen, X.~Shen, J.~Hao, and J.~Wang.
\newblock Causal{VAE}: Structured causal disentanglement in variational
  autoencoder.
\newblock \emph{arXiv preprint arXiv:2004.08697}, 2020.

\bibitem[Yang et~al.(2018)Yang, Hu, Dyer, Xing, and
  Berg{-}Kirkpatrick]{yang2018unsupervised}
Z.~Yang, Z.~Hu, C.~Dyer, E.~P. Xing, and T.~Berg{-}Kirkpatrick.
\newblock Unsupervised text style transfer using language models as
  discriminators.
\newblock In \emph{NeurIPS}, pages 7298--7309, 2018.

\bibitem[Yao et~al.(2020)Yao, Chu, Li, Li, Gao, and Zhang]{yao2020survey}
L.~Yao, Z.~Chu, S.~Li, Y.~Li, J.~Gao, and A.~Zhang.
\newblock A survey on causal inference.
\newblock \emph{arXiv preprint arXiv:2002.02770}, 2020.

\bibitem[Zellers et~al.(2020)Zellers, Holtzman, Rashkin, Bisk, Farhadi,
  Roesner, and Choi]{zellers2020defending}
R.~Zellers, A.~Holtzman, H.~Rashkin, Y.~Bisk, A.~Farhadi, F.~Roesner, and
  Y.~Choi.
\newblock Defending against neural fake news.
\newblock \emph{Neurips}, 2020.

\bibitem[Zeng et~al.(2020)Zeng, Li, Zhai, and Zhang]{zeng2020counterfactual}
X.~Zeng, Y.~Li, Y.~Zhai, and Y.~Zhang.
\newblock Counterfactual generator: A weakly-supervised method for named entity
  recognition.
\newblock In \emph{Proceedings of the 2020 Conference on Empirical Methods in
  Natural Language Processing (EMNLP)}, pages 7270--7280, 2020.

\bibitem[Zhang et~al.(2020)Zhang, Zhang, Tang, Hua, and Sun]{zhang2020causal}
D.~Zhang, H.~Zhang, J.~Tang, X.-S. Hua, and Q.~Sun.
\newblock Causal intervention for weakly-supervised semantic segmentation.
\newblock \emph{Advances in Neural Information Processing Systems}, 33, 2020.

\bibitem[Zhao et~al.(2017)Zhao, Wang, Yatskar, Ordonez, and Chang]{zhao2017men}
J.~Zhao, T.~Wang, M.~Yatskar, V.~Ordonez, and K.-W. Chang.
\newblock Men also like shopping: Reducing gender bias amplification using
  corpus-level constraints.
\newblock In \emph{Proceedings of the 2017 Conference on Empirical Methods in
  Natural Language Processing}, 2017.

\bibitem[Zhu et~al.(2020)Zhu, Zhang, Liu, and Wang]{zhu2020counterfactual}
Q.~Zhu, W.~Zhang, T.~Liu, and W.~Y. Wang.
\newblock Counterfactual off-policy training for neural dialogue generation.
\newblock In \emph{Proceedings of the 2020 Conference on Empirical Methods in
  Natural Language Processing (EMNLP)}, pages 3438--3448, 2020.

\bibitem[Zmigrod et~al.(2019)Zmigrod, Mielke, Wallach, and
  Cotterell]{zmigrod2019counterfactual}
R.~Zmigrod, S.~J. Mielke, H.~Wallach, and R.~Cotterell.
\newblock Counterfactual data augmentation for mitigating gender stereotypes in
  languages with rich morphology.
\newblock In \emph{Proceedings of the 57th Annual Meeting of the Association
  for Computational Linguistics}, pages 1651--1661, 2019.

\end{thebibliography}

\end{document}

%\maketitle

\appendix

\section{Latent Confounders and Proxy}

It is infeasible to estimate causal effects without any observed confounding information \citep{pearl2009causality,d2019multi}. With unobserved confounders, a common approach is to introduce additional proxy variables \citep{montgomery2000measuring,angrist2008mostly,stock2012introduction}. For example, the socio-economic status of a patient is a confounder to the medication and outcome, which is unobserved and hard to measure directly. But we can use known auxiliary information such as the zip code and job type as a proxy to assess the confounder indirectly \citep{louizos2017causal}. Note that directly treating the proxy variables as ordinary confounders can induce bias \citep{pearl2012measurement,griliches1986errors,fuller2009measurement}. Instead, recent work \citep{louizos2017causal,tran2018implicit,lu2018considering,madras2019fairness,ranganath2018multiple} uses latent variable approaches. For example, \citet{louizos2017causal} use VAEs to infer latent confounders from proxy information. Our work develops a latent variable causal framework for controllable text generation. Due to the rich and abstract information in text, we introduce necessary training objectives to avoid training collapse and encourage confounder balance.

\section{Experimental Details}

\paragraph{Model configuration}
The VAE backbone of our model largely follows the architecture of the VAE model in \cite{li2020optimus}, except that our inference network (encoder) $q_\phi$ adopts the pretrained GPT-2 model (same as the decoder $p_\theta(\x|a,\z)$) instead of BERT, in order to make sure both encoder $q_\phi$ and decoder $p_\theta(\x|a,\z)$ have the same tokenization. We implement $a$ to be either an all-zero or all-one vector of dimension $50$, and set the dimension of $\z$ to $718$, so that concatenating $a$ and $\z$ leads to a vector of dimension $768$, same as in \cite{li2020optimus}. We implement $p_\theta(\bc|\z)$ as a single-layer MLP and $p_\theta(a|\z)$ as a two-layer MLP with the intermediate dimension the same as the input dimension ($768$).

\paragraph{More details of \textsc{Bios} dataset}

For occupation which is the confounding factor, we subsample and merge the occupations into two groups, i.e., \emph{\{nurse, dietitian, paralegal, model, yoga teacher\}}, and \emph{\{rapper, DJ, surgeon, software engineer, composer\}}, which results in a correlation strength of $94\%$ between occupation and gender.

\paragraph{Training confounding label classifier with data re-weighting}
For the \texttt{Conditional LM (full)} baseline for attribute-conditional generation, we first train a confounding label classifier with the limited confounding (proxy) labels in the dataset. Due to the strong correlation between the confounding factor and attribute (e.g., with a correlation strength of 90\%), only a small fraction of (e.g., 10\%) instances have opposite confounding label and attribute label. We thus train the classifier with data reweighting to reduce the bias. Specifically, we associate a weight of 0.9 to those instances with opposite confounding and attribute labels, and a weight of 0.1 to other instances with the same confounding and attribute labels. We tried other weights and obtained lower or similar classifier accuracy.

\paragraph{Human evaluation of text attribute transfer}

Following previous work \citep[e.g.,][]{lin2020record}, we conduct comparison-based human evaluation for the output of different generation models. Specifically, for each test instance, we present the outputs of two comparison models to the human rater, and ask the human rater to rank which of the two outputs are better in terms of the goal of the task (i.e., accurately rewriting the text to possess desired attribute and meanwhile preserving all other characteristics of the original sentence). The human rater can also choose “no preference” if the two outputs are equally good or bad. We asked three human annotators (who are graduate students and proficient English speakers) to do the rating \citep{lin2020record}. There were no potential participant risks.
We evaluate on 50 test cases for each pair of comparison models. Table~\ref{tab:tst-biased-human} shows the results, which are consistent with the observations from automatic evaluation. 

\begin{table*}[h]
%\vspace{-15pt}
\centering
\small
\begin{tabular}{r c c c}
\cmidrule[\heavyrulewidth](lr){1-4}
& Ours better (\%) & No preference (\%)  & Ours worse (\%)   \\
\cmidrule[\heavyrulewidth](lr){1-4}
\citet{Hu2017TowardCG} &  {\bf 62} & 24  & 14  \\
Ablation: {Ours} w/o ${cf\text{-}z/c}$ & {\bf 54} & 22 & 24  \\
\cmidrule[\heavyrulewidth](lr){1-4} 
\end{tabular}
\vspace{-4pt}
\caption{
Human evaluation of text attribute transfer on biased \textsc{Yelp}.
For example, the outputs of \textsc{Ours} are considered to be better than those of \citet{Hu2017TowardCG} on 62\% test instances.
}
\label{tab:tst-biased-human}
%\vspace{-16pt}
\end{table*}

\paragraph{Classifiers used in training and evaluation}
We summarize the different classifiers used in training and evaluation in the experiment to serve as a reference and avoid confusion.
\begin{itemize}
    \item For training:
    \begin{itemize}
        \item \emph{attribute classifier} $f$ (Eq.4) is used to train our causal model. The classifier is pretrained with the biased (attribute, text) training corpus.
        \item \emph{confounding label classifier} is used in the baseline \texttt{Conditional LM (full)} for attribute-conditional generation. The classifier is trained with the available confounding (proxy) labels with data re-weighting, as discussed above.
    \end{itemize}
    \item For evaluation:
    \begin{itemize}
        \item \emph{evaluation attribute classifier} is used to evaluate the generation accuracy of desired attribute. As an evaluation metric, the classifier is obtained by training on additional \emph{unbiased} (attribute, text) data.
        \item \emph{evaluation confounding label classifier} is used to evaluate the correlation of attribute and confounding (proxy) labels in the generation. Similarly, the classifier is obtained by training on additional large unbiased data.
    \end{itemize}
\end{itemize}

\section{Generated Samples}

\begin{table*}[h]
%\vspace{-15pt}
\centering
%\small
\setlength{\tabcolsep}{8pt} % Default value: 6pt
\renewcommand{\arraystretch}{1.3} % Default value: 1
\begin{tabular}{l}
\cmidrule[\heavyrulewidth](lr){1-1}
\textsc{Conditional LM (full)} \\
\cmidrule(lr){1-1}
{\bf $a=0$ (sentiment negative)} \\
\texttt{this was the worst experience i 've ever had at a glazier .}  \\
\texttt{i even asked him if they could play on the tv channel . } \\
\texttt{this was pretty fun the first time i went .  "} \\
\texttt{waited in line once but almost never reached the floor .} \\
\texttt{if you are ever up in chandler , tony will stop by .} \\
{\bf $a=1$ (sentiment positive)} \\
\texttt{very good and long wait time . }\\
\texttt{we loved our favorite harrah 's night ! "} \\
\texttt{i would love to try this restaurant again when they open .  "} \\
\texttt{this place is great .} \\
\texttt{everything you will find in this restaurant !} \\
\cmidrule[\heavyrulewidth](lr){1-1} 
\textsc{Ours} \\
\cmidrule(lr){1-1} 
{\bf $a=0$ (sentiment negative)} \\
\texttt{no , it 's obvious that they were overcooked .} \\
\texttt{the seats were poorly done and basically sucked up .} \\
\texttt{it was n't enough to ask us if it was okay .} \\
\texttt{very disappointed with my food order yesterday .} \\
\texttt{i declined to replace it tho they were bad .} \\
{\bf $a=1$ (sentiment positive)} \\
\texttt{great for a relaxed evening out .} \\
\texttt{i 'm beyond impressed with the passion fruit and unbeatable service .} \\
\texttt{it 's a true pleasure to meet andrew .} \\
\texttt{jacksville became my go-to spot for dessert .}\\
\texttt{thank you for the technique , i am quite impressed .}\\
\cmidrule[\heavyrulewidth](lr){1-1} 
\end{tabular}
\vspace{-4pt}
\caption{Attribute-conditional generation trained on \textsc{Yelp} dataset. \textsc{Conditional LM (full)} tends to generate non-restaurant reviews conditioning on $a=0$, and restaurant reviews conditioning on $a=1$.}
\label{tab:samples-cond-gen-yelp}
%\vspace{-16pt}
\end{table*}

\begin{table*}[h]
%\vspace{-15pt}
\centering
\small
\setlength{\tabcolsep}{8pt} % Default value: 6pt
\renewcommand{\arraystretch}{1.3} % Default value: 1
\begin{tabular}{l}
\cmidrule[\heavyrulewidth](lr){1-1}
\textsc{Ours} \\
\cmidrule(lr){1-1} 
{\bf $a=0$ (sentiment negative) $\to$ $a=1$ (positive)} \\
original: \texttt{pick-up was just ok , but vehicle was filthy and had trash in it .} \\
output:\quad \texttt{pick-up was pretty good , but atmosphere was just incredible and comfortable .} \\
\\
original: \texttt{so that was nice but they served some sweet concoctions that made me sick .} \\
output:\quad \texttt{so good that they served some sweet and flavorful cocktails that made me super happy .} \\
\\
original: \texttt{similar to some of the other reviewers , the poutine was just that .} \\
output:\quad \texttt{similar to some of the other reviewers , the poutine was just perfect .}\\
\\
{\bf $a=1$ (sentiment positive) $\to$ $a=0$ (negative)} \\
original: \texttt{the santa fe salad is awesome .}\\
output:\quad \texttt{the santa fe salad is mediocre .} \\
\\
original: \texttt{the employees were super helpful , friendly and attentive .}\\
output:\quad \texttt{the employees were super rude , incompetent and unhelpful .} \\
\\
original: \texttt{i love their eggs benedict and pancakes both are amazing !}\\
output:\quad \texttt{i hate their eggs benedict and pancakes both are horrible .}\\
\cmidrule[\heavyrulewidth](lr){1-1} 
\end{tabular}
\vspace{-4pt}
\caption{Text attribute transfer on the biased \textsc{Yelp} dataset.}
\label{tab:samples-cond-gen-yelp}
%\vspace{-16pt}
\end{table*}

\bibliographystyle{abbrvnat}
\bibliography{custom}